\documentclass[12pt]{article}

\usepackage{natbib}
\usepackage{amsmath,amssymb}
\usepackage{graphicx}
\usepackage{booktabs}
\usepackage{multirow}
\usepackage{float}
\usepackage{xcolor}
\usepackage{textcomp}
\usepackage[hypertexnames=false]{hyperref}

\graphicspath{{figures/}}

\title{Physical Knot Classification Beyond Accuracy: A Benchmark and Diagnostic Study}

\author{Shiheng Nie \and Yunguang Yue\thanks{Corresponding author. \texttt{guangyy@shzu.edu.cn}}\\
College of Science, Shihezi University, Shihezi 832003, Xinjiang, China}

\date{}

\begin{document}

\maketitle

\begin{abstract}
Physical knot classification is a challenging fine-grained recognition task in which the intended discriminative cue is rope crossing structure; however, high closed-set accuracy may still arise from low-level appearance shortcuts rather than genuine topological understanding. In this work, we introduce Knots-10, a tightness-stratified benchmark constructed from the public 10Knots dataset (1{,}440 images, 10 classes), which trains models on loosely tied knots and evaluates them on tightly dressed configurations to probe whether structure-guided training yields topology-specific gains. We demonstrate that topological distance successfully predicts residual inter-class confusion across multiple backbone architectures, validating the utility of our topology-aware evaluation framework. Furthermore, we propose topology-aware centroid alignment (TACA) and an auxiliary crossing-number prediction objective as two complementary forms of structural supervision. Notably, Swin-T with TACA achieves a consistent positive specificity gain ($\Delta_{\text{spec}} = +1.18$~pp) across all random seeds under the canonical protocol, and auxiliary crossing-number prediction exhibits robust performance across data regimes without the real-versus-random reversal observed for centroid alignment. Causal probes reveal that background changes alone flip 17--32\% of predictions and phone-photo accuracy drops by 58--69 percentage points, underscoring that appearance bias remains the principal obstacle to deployment. These results collectively demonstrate that our diagnostic workflow provides a principled and practical tool for evaluating whether a hand-crafted structural prior delivers genuine task-relevant benefit beyond generic regularization.
\end{abstract}

\noindent\textbf{Keywords:} knot classification; fine-grained recognition; benchmark dataset; structure-guided learning; appearance bias

\section{Introduction}
\label{sec:intro}

Physical knot classification is a constrained fine-grained recognition problem \citep{wei2022fgia} in which different classes are tied with the same rope under similar capture conditions, causing low-level appearance cues to be broadly shared across categories. The intended discriminative signal is crossing structure, making knot images a uniquely informative testbed for investigating whether high-accuracy models genuinely rely on structural cues or exploit easier appearance shortcuts.

The problem carries practical significance for climbing and surgical safety \citep{adams2004, zimmer2020surgical}, and related topological reasoning underlies emerging applications in DNA sensing and knotted protein analysis \citep{sharma2019complexdna, hsu2023knottedproteins}. Despite its relevance, the public 10Knots dataset \citep{cameron2018tenknots} lacks a standardized evaluation protocol and provides little diagnostic evidence about the cues models actually exploit. To address this gap, we introduce Knots-10, a tightness-stratified benchmark that trains on loosely tied knots and evaluates on tightly dressed configurations, thereby simulating realistic deployment conditions.

\begin{figure}[H]
\centering
\includegraphics[width=0.97\linewidth]{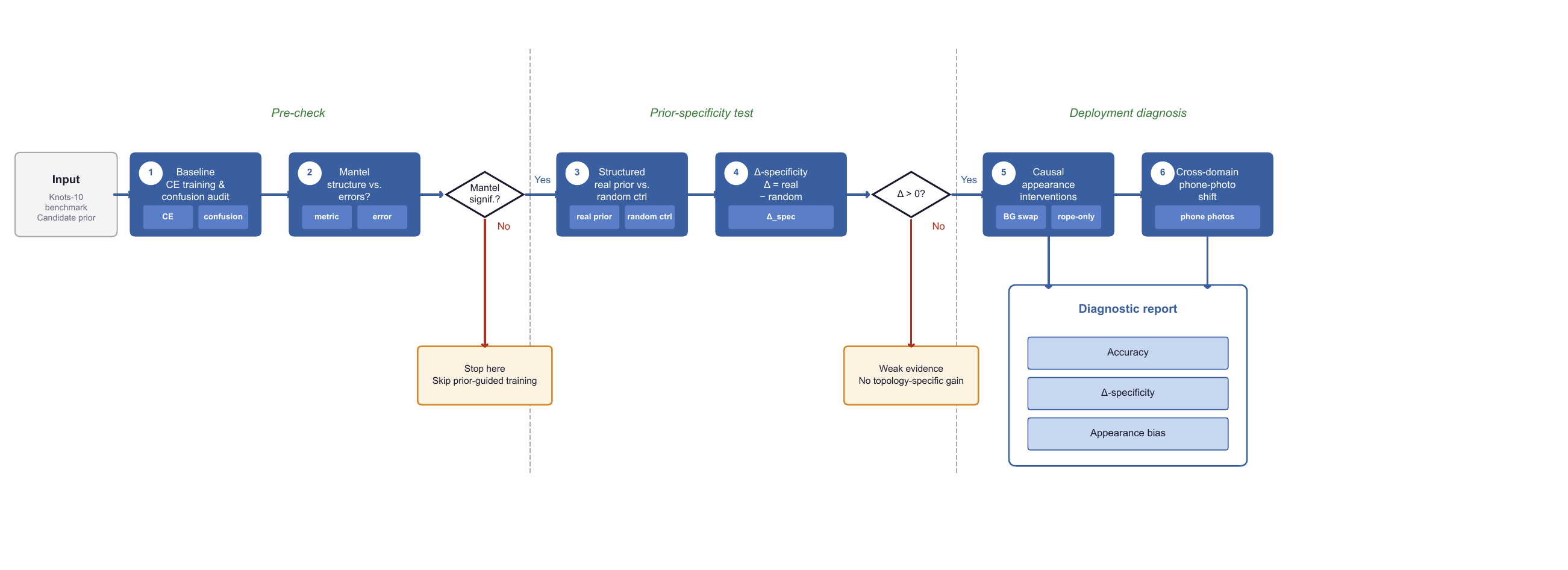}
\caption{\textbf{Diagnostic workflow for prior-specific evaluation on Knots-10.} Baseline training and a Mantel pre-check first decide whether a candidate structural prior is worth testing. If the Mantel signal is significant, the workflow compares real-prior and random-prior training through $\Delta_{\text{spec}}$ and only then proceeds to causal probes and cross-domain tests.}
\label{fig:overview}
\end{figure}

We employ random-distance controls, Mantel correlation analysis, causal appearance probes, and cross-domain evaluation to rigorously assess whether apparent gains from structure-guided training are genuinely prior-specific. Our analysis demonstrates that topological distance reliably predicts inter-class confusion patterns, and that topology-aware centroid alignment (TACA) achieves consistent specificity gains for strong transformer backbones. Figure~\ref{fig:overview} illustrates the diagnostic workflow used to determine whether a candidate structural prior yields prior-specific benefit beyond generic regularization. The main contributions of this work are summarized as follows:

\begin{enumerate}
    \item We introduce Knots-10, a tightness-stratified benchmark constructed from the public 10Knots dataset, providing a rigorous protocol for evaluating structure-aware recognition from loosely tied to tightly dressed knot configurations.

    \item We propose a diagnostic workflow based on $\Delta_{\text{spec}}$, random-distance controls, causal appearance probes, and domain generalization (DG) composition tests, enabling principled evaluation of structural priors.

    \item We demonstrate that topological distance predicts inter-class confusion across multiple architectures, and that TACA achieves consistent topology-specific gains for Swin-T, while auxiliary crossing-number prediction offers a more stable structural supervision signal across data regimes.
\end{enumerate}

\section{Related Work}
\label{sec:related}

The majority of learning-based work on knots employs mathematical rather than photographic inputs, such as planar diagrams or polymer coordinate sequences \citep{hughes2020neural, vandans2020knots}. While robotics studies have utilized vision for rope manipulation tasks \citep{sundaresan2020}, fine-grained knot recognition from photographs remains largely unexplored. Our setting is therefore distinct in both input modality and evaluation objective.

Physical knot classification is also a fine-grained visual classification (FGVC) task \citep{wei2022fgia}. Recent FGVC methods have achieved notable advances through part selection, local relation modeling, and feature enhancement \citep{he2022transfg, wang2023aatrans, zhuang2020apinet, du2020pmg, chen2024fetfgvc, shi2025ldhvit, ke2025multigranularity}; however, these methods predominantly employ generic visual priors and do not investigate whether a domain-specific structural prior survives random-prior controls.

The tightness-stratified split additionally constitutes a domain generalization (DG) setting, as training and test images differ systematically in knot dressing and, in the phone-photo pilot, in rope appearance and scene conditions \citep{zhou2022domain}. Prior DG work has improved robustness through style or frequency perturbations \citep{xu2023fourier, hu2024dg_alignment}; nevertheless, recent analyses demonstrate that such gains depend strongly on backbone architecture and data regime \citep{backbones2024, smalldatadg2024}. More broadly, our work relates to structured-prior learning \citep{barz2019hierarchy, palazzo2021structured}, with a particular emphasis on rigorously verifying whether a hand-crafted prior delivers genuine task-relevant benefit.

\section{Materials and Methods}\label{sec:methods}

\subsection{Dataset and Evaluation Protocol}\label{sec:dataset}

\subsubsection{Dataset}

We employ the public \textbf{10Knots} dataset \citep{cameron2018tenknots}, released under a CC BY-SA 4.0 license, which comprises 1{,}440 photographs of 10 knot types tied with climbing rope. Table~\ref{tab:knot_taxonomy} summarizes the class topology utilized in our topological distance metric.

\begin{table}[t]
\centering
\caption{Topological taxonomy of Knots-10. $C_{\text{vis}}$ denotes the visual crossing number (see Remark below). These properties define the topological distance metric in Section~\ref{sec:topo_metric}.}
\label{tab:knot_taxonomy}
\small
\begin{tabular}{@{}lllll@{}}
\toprule
ID & Knot Name & Type & $C_{\text{vis}}$ & Family \\
\midrule
OHK & Overhand Knot & Prime ($3_1$) & 3 & Stopper \\
SK  & Slip Knot & Slip ($3_1$) & 3 & Stopper \\
F8K & Figure-8 Knot & Prime ($4_1$) & 4 & Stopper \\
BK  & Bowline & Loop & 4 & Loop \\
F8L & Figure-8 Loop & Loop ($4_1$) & 4 & Loop \\
ABK & Alpine Butterfly & Loop & 4 & Loop \\
CH  & Clove Hitch & Hitch & 2 & Hitch \\
RK  & Reef Knot & Composite ($3_1 \# 3_1$) & 6 & Binding \\
FSK & Fisherman's Knot & Composite ($3_1 \# 3_1$) & 6 & Bend \\
FMB & Flemish Bend & Composite ($4_1 \# 4_1$) & 8 & Bend \\
\bottomrule
\end{tabular}
\end{table}

\noindent\textbf{Remark.} Only four classes in our set correspond to formal mathematical knots: Overhand Knot (OHK), Figure-8 Knot (F8K), Reef Knot (RK), and Fisherman's Knot (FSK). The remaining classes are practical rope configurations whose $C_{\text{vis}}$ values are visual estimates. The topological distance metric (Section~\ref{sec:topo_metric}) is therefore a knot-theory-inspired heuristic (see Section~\ref{sec:discussion}).

Each class contains 144 images from 3 lighting conditions $\times$ 3 tightness levels $\times$ 16 instances \citep{cameron2018tenknots}. All images show climbing rope from a top-down perspective.

We adopt a \emph{tightness-stratified} partition, training on Loose and VeryLoose images (768 samples after a 20\% validation holdout) and evaluating on all tightly dressed Set images (480 samples). This protocol simulates realistic deployment conditions, where knots are encountered in a tightly dressed state while training data originates from loosely tied demonstrations.

\subsection{Topological Distance}
\label{sec:topo_metric}

We represent inter-class similarity through a principled hand-crafted pairwise topological distance $d(k_i, k_j)$ between knot classes:

\begin{equation}
d(k_i, k_j) = \sum_{f=1}^{5} w_f \cdot \delta_f(k_i, k_j)
\end{equation}

\begin{sloppypar}
\noindent where the five factors are:
$\delta_1$~= normalized crossing-number difference ($|C_{\text{vis}}(k_i) - C_{\text{vis}}(k_j)|/8$),
$\delta_2$~= family membership~(0/1),
$\delta_3$~= type similarity~(0/0.5),
$\delta_4$~= component count difference, and
$\delta_5$~= structural derivation penalty encoding shared base structure (e.g., Overhand--Slip = 0.1; full matrix in Supplementary Table~S4).
Weights are $w_{1,2,5} = 0.25$, $w_3 = 0.15$, $w_4 = 0.10$. Both factors and weights are author-assigned heuristics informed by knot theory \citep{adams2004}; Supplementary Figure~S4 shows robustness to $\pm 50\%$ perturbations.
\end{sloppypar}

\subsection{Models}
\label{sec:models}

We evaluate five general-purpose backbones spanning convolutional neural networks (CNNs) and vision transformers: ResNet-18 (11.2M parameters) \citep{he2016resnet}, ResNet-50 (23.5M), EfficientNet-B0 (4.0M) \citep{tan2019efficientnet}, ViT-B/16 (85.8M) \citep{dosovitskiy2021vit}, and Swin-T (27.5M) \citep{liu2021swin}. Additionally, we evaluate three FGVC-specialized architectures: TransFG \citep{he2022transfg}, PMG \citep{du2020pmg}, and Graph-FGVC, a graph-reasoning variant inspired by API-Net with graph convolutional network (GCN) layers \citep{zhuang2020apinet, kipf2017gcn}. All models are initialized with ImageNet-pretrained weights \citep{russakovsky2015imagenet} and fine-tuned end-to-end.

\subsection{Training Setup}
\label{sec:training}

All models are trained for 20 epochs using the AdamW optimizer \citep{loshchilov2019adamw} with a learning rate of $10^{-4}$, weight decay of $10^{-4}$, cosine annealing schedule, input resolution $224 \times 224$, batch size 32, and standard data augmentation comprising horizontal flipping, $\pm 15^{\circ}$ rotation, and light color jitter. The checkpoint achieving the highest validation accuracy is selected for test evaluation.

All experiments are conducted in PyTorch 2.1 on NVIDIA A800 GPUs. Main-paper results report mean $\pm$ standard deviation across three random seeds (42, 123, 456); per-seed breakdowns are provided in Supplementary Table~S1.

\subsection{Structure-Guided Objectives}
\label{sec:topo_method}

We propose two forms of structure-guided supervision augmenting the standard cross-entropy loss. The objective is to investigate whether weak structural supervision encourages learned representations to respect knot topological similarity. Let $\mathbf{z}_i$ denote the penultimate-layer representation for sample $i$, and let $\boldsymbol{\mu}_k = \frac{1}{|S_k|}\sum_{i \in S_k} \mathbf{z}_i$ be the class centroid for class $k$ within a mini-batch $S_k$.

\subsubsection{Topology-Aware Centroid Alignment (TACA)}

The topology-aware centroid alignment (TACA) loss encourages the pairwise distances between class centroids in the embedding space to align with the topological distance matrix, thereby injecting structural inductive bias into the learned representations:

\begin{equation}
\mathcal{L}_{\text{TACA}} = \frac{1}{K^2}\left\|\hat{\mathbf{D}}^{\text{emb}} - \mathbf{D}^{\text{topo}}\right\|_F^2
\end{equation}
where $\hat{\mathbf{D}}^{\text{emb}}_{k,k'} = \|\boldsymbol{\mu}_k - \boldsymbol{\mu}_{k'}\|_2 / \max_{i,j}\|\boldsymbol{\mu}_i - \boldsymbol{\mu}_j\|_2$ is the max-normalized embedding distance. Classes absent from a mini-batch are omitted from the loss for that batch.
Additional TACA variants, including a contrastive form (TAML) and learnable distance weights, are reported only in the Supplementary Material.

\subsubsection{Auxiliary Crossing Prediction}
\label{sec:auxcrossing}

To investigate whether explicit crossing-number supervision enhances classification performance, we introduce an auxiliary prediction head that estimates the visual crossing number $C_{\text{vis}} \in \{2, 3, 4, 6, 8\}$, corresponding to the five unique values in Table~\ref{tab:knot_taxonomy}. The head comprises a linear layer trained with cross-entropy loss:
\begin{equation}
\mathcal{L}_{\text{aux}} = \text{CE}(\text{softmax}(\mathbf{W}_{\text{aux}} \mathbf{z}_i), \, c_{\text{vis}})
\end{equation}
where $c_{\text{vis}}$ is the bin index corresponding to the sample's $C_{\text{vis}}$. The combined loss is $\mathcal{L} = \mathcal{L}_{\text{CE}} + \lambda_{\text{aux}} \cdot \mathcal{L}_{\text{aux}}$ with $\lambda_{\text{aux}} = 0.1$. This approach tests whether coarse topological structure (crossing count) provides useful inductive bias, without requiring the full pairwise distance matrix used in TACA.

For DG comparison we use MixStyle \citep{zhou2021mixstyle} and AmpMix \citep{wang2021ampmix}. Statistical significance is assessed with the Mantel permutation test \citep{mantel1967} for topology--confusion correlation and McNemar's test \citep{mcnemar1947} for paired classifier comparisons; details are given in Supplementary Section~S6.

\section{Experiments}\label{sec:results}

\subsection{Baseline Results}\label{sec:baseline}

We first evaluate general-purpose backbones and FGVC-specialized models on Knots-10. Swin-T achieves the highest accuracy among general architectures at $97.2 \pm 1.1\%$ (Table~\ref{tab:results}), with the remaining four models clustered within 2 percentage points.

\begin{table}[t]
\centering
\caption{\textbf{Classification results on Knots-10} (mean $\pm$ std across seeds 42, 123, and 456). All models use ImageNet pretraining and the tightness-stratified protocol.}
\label{tab:results}
\small
\begin{tabular}{@{}lccc@{}}
\toprule
Model & Params & Test Acc (\%) & Macro F1 \\
\midrule
\multicolumn{4}{@{}l}{\textit{General Architectures}} \\
\addlinespace[2pt]
ResNet-18 & 11.2M & $96.88 \pm 1.06$ & $0.969 \pm 0.011$ \\
ResNet-50 & 23.5M & $95.83 \pm 1.03$ & $0.959 \pm 0.010$ \\
EfficientNet-B0 & 4.0M & $96.25 \pm 0.45$ & $0.963 \pm 0.005$ \\
ViT-B/16 & 85.8M & $96.39 \pm 0.35$ & $0.964 \pm 0.003$ \\
\textbf{Swin-T} & \textbf{27.5M} & $\mathbf{97.22 \pm 1.09}$ & $\mathbf{0.972 \pm 0.011}$ \\
\midrule
\multicolumn{4}{@{}l}{\textit{FGVC-Specialized Methods}} \\
\addlinespace[2pt]
TransFG & 86.4M & $97.15 \pm 0.94$ & $0.972 \pm 0.010$ \\
PMG & 23.6M & $94.51 \pm 1.75$ & $0.945 \pm 0.018$ \\
Graph-FGVC & 26.8M & $95.49 \pm 2.84$ & $0.955 \pm 0.028$ \\
\bottomrule
\end{tabular}
\end{table}

McNemar paired tests (Supplementary Section~S6) demonstrate that Swin-T is the only general-purpose model significantly superior to all others ($p < 10^{-4}$), while the remaining four are statistically equivalent ($p > 0.57$). Among FGVC-specialized methods, TransFG matches Swin-T at $97.2 \pm 0.9\%$, whereas PMG ($94.5 \pm 1.8\%$) and Graph-FGVC ($95.5 \pm 2.8\%$) yield lower performance. Differences below approximately 2 percentage points fall within sampling uncertainty at this test set size.

Per-class and lighting condition breakdowns follow consistent patterns across architectures. Clove Hitch is the most reliably classified category, Figure-8 Knot presents the greatest challenge, and the Reef Knot--Fisherman's Knot pair exhibits pronounced sensitivity to lighting variation; comprehensive breakdowns are provided in Supplementary Figure~S2.

\subsection{Topological Correlation}\label{sec:correlation}

We next investigate whether topological similarity predicts inter-class confusion patterns. Specifically, if the topological distance metric (Section~\ref{sec:topo_metric}) captures genuine structural relationships, it should align with observed confusion rates across architectures.

Spearman and Pearson correlations between topological distance and pairwise confusion rate across all 45 knot pairs are reported in Table~\ref{tab:correlation}.

\begin{table}[t]
\centering
\caption{Correlation between topological distance and pairwise confusion rate. Mantel $p$-values use 9{,}999 permutations to account for matrix non-independence \citep{mantel1967}.}
\label{tab:correlation}
\small
\begin{tabular}{@{}lcccc@{}}
\toprule
Model & Spearman $\rho$ & Pearson $r$ & Mantel $r$ & Mantel $p$ \\
\midrule
ResNet-18 & $-0.291$ & $-0.301$ & $-0.291$ & 0.058 \\
ResNet-50 & $-0.472$ & $-0.453$ & $-0.472$ & 0.002\rlap{$^{**}$} \\
EfficientNet-B0 & $-0.491$ & $-0.426$ & $-0.492$ & 0.001\rlap{$^{***}$} \\
ViT-B/16 & $-0.398$ & $-0.359$ & $-0.398$ & 0.005\rlap{$^{**}$} \\
Swin-T & $-0.214$ & $-0.195$ & $-0.214$ & 0.173 \\
\bottomrule
\multicolumn{5}{@{}l}{\footnotesize $^{**}p<0.01$; $^{***}p<0.001$. 9{,}999 permutations, two-tailed.}
\end{tabular}
\end{table}

Three of five architectures yield statistically significant negative Mantel correlations, confirming that topologically similar knot pairs are more frequently confused. EfficientNet-B0 exhibits the strongest topology--confusion alignment ($r = -0.492$, $p = 0.001$), with ResNet-50 and ViT-B/16 also achieving significance (both $p < 0.01$). Notably, Swin-T does not reach significance ($p = 0.173$), attributable to its sparse confusion pattern (only 3 non-zero off-diagonal pairs), which limits statistical power rather than indicating an absence of topological structure.

It is worth noting that topology--confusion alignment does not increase monotonically with model capacity. EfficientNet-B0, a mid-capacity model, demonstrates the strongest alignment, and the most frequently confused pairs correspond precisely to topologically proximate classes: Figure-8 Knot--Overhand Knot ($d = 0.156$), Figure-8 Knot--Slip Knot ($d = 0.231$), and Fisherman's Knot--Flemish Bend ($d = 0.150$). Supplementary Figures~S3--S5 further demonstrate that these correlations are robust to moderate metric weight perturbations.

\subsection{Canonical TACA Ablation}\label{sec:topo_results}

We then test whether a topology-aware loss improves learning beyond generic regularization. Under the canonical protocol, Swin-T showed a positive $\Delta_{\text{spec}}$ across three seeds, while ViT-B/16 and ResNet-50 showed mixed or near-zero responses. This signal was not stable across protocols. In the reduced-data experiment (Section~\ref{sec:reduced_data}), a different TACA normalization gives $\Delta_{\text{spec}} \approx 0$. We therefore do not treat the Swin-T result as a general topology-specific effect.

We investigate whether TACA provides topology-specific improvement beyond generic regularization by comparing three training conditions across ResNet-50, Swin-T, and ViT-B/16 (Table~\ref{tab:taca_arch}): (1)~cross-entropy (CE) only, (2)~CE + TACA with the real topological distance matrix, and (3)~CE + TACA with a randomly permuted distance matrix. All models are trained for 20 epochs on the canonical split, and results report mean $\pm$ std across seeds 42, 123, and 456. AuxCrossing is discussed separately in Sections~\ref{sec:auxcrossing} and \ref{sec:dg_baselines}.

\begin{table}[t]
\centering
\caption{\textbf{TACA ablation under the canonical protocol} (mean $\pm$ std across seeds 42, 123, and 456; $n = 480$). $\Delta_{\text{spec}}$ = TACA(real) $-$ TACA(rand). For protocol sensitivity, compare with the reduced-data suite in Table~\ref{tab:reduced_data}.}
\label{tab:taca_arch}
\small
\begin{tabular}{@{}lccc@{}}
\toprule
Condition & ResNet-50 & Swin-T & ViT-B/16 \\
\midrule
CE only & $95.76 \pm 0.71$ & $97.15 \pm 0.79$ & $97.29 \pm 0.17$ \\
TACA (rand) & $96.46 \pm 0.78$ & $97.64 \pm 0.35$ & $95.56 \pm 2.35$ \\
TACA (real) & $96.18 \pm 1.58$ & $\mathbf{98.82 \pm 0.26}$ & $96.46 \pm 1.06$ \\
\midrule
$\Delta_{\text{spec}}$ (pp) & $-0.28 \pm 1.82$ & $\mathbf{+1.18 \pm 0.20}$ & $+0.90 \pm 1.97$ \\
\bottomrule
\multicolumn{4}{@{}l}{\footnotesize All values in \%. Per-seed $\Delta_{\text{spec}}$: Swin-T ($+$1.05, $+$1.04, $+$1.46),}\\
\multicolumn{4}{@{}l}{\footnotesize ViT-B/16 ($+$3.13, $+$1.25, $-$1.67), ResNet-50 ($-$2.71, $+$0.21, $+$1.67).}
\end{tabular}
\end{table}

Under the canonical protocol, \textbf{Swin-T} with TACA achieves $98.82 \pm 0.26\%$, yielding a consistent positive specificity gain of $\Delta_{\text{spec}} = +1.18 \pm 0.20$~pp across all three random seeds, which is the most stable topology-specific signal observed across all architectures. \textbf{ViT-B/16} also shows a positive mean $\Delta_{\text{spec}}$ of $+0.90$~pp, though with higher variance across seeds. \textbf{ResNet-50} shows near-zero $\Delta_{\text{spec}}$, suggesting that CNN-based architectures may require stronger inductive capacity to benefit from centroid-level topological alignment. Per-seed values are listed in Table~\ref{tab:taca_arch}. Exploratory pair-level analysis and additional CNN results are provided in Supplementary Sections~S14--S15. Taken together, these results indicate that the Mantel test serves as an effective go/no-go screening criterion before applying distance-guided regularization.

\subsection{Reduced-Data Stress Test}
\label{sec:reduced_data}

We next investigate whether structural supervision provides greater benefit under limited data conditions. Swin-T is trained under four conditions: CE baseline, TACA with real topological distances, TACA with randomly permuted distances, and auxiliary crossing-number prediction with ordinal encoding (AuxCrossing-Ord). Results are reported in Table~\ref{tab:reduced_data}.

\begin{table}[t]
\centering
\caption{\textbf{Reduced-data stress test} for Swin-T (mean $\pm$ std across seeds 42, 123, and 456). This is an independent experimental suite with a different TACA normalization, so only within-table comparisons are valid. $\Delta_{\text{spec}}$ = TACA(real) $-$ TACA(rand).}
\label{tab:reduced_data}
\small
\begin{tabular}{@{}lcc@{}}
\toprule
Method & 50\% data & 100\% data \\
\midrule
CE (baseline) & $94.31 \pm 1.58$ & $97.15 \pm 0.96$ \\
TACA (real) & $96.67 \pm 0.63$ & $96.67 \pm 0.42$ \\
TACA (rand) & $96.60 \pm 0.43$ & $98.19 \pm 0.73$ \\
AuxCrossing-Ord & $96.60 \pm 0.97$ & $97.08 \pm 0.42$ \\
\midrule
$\Delta_{\text{spec}}$ (pp) & $+0.07$ & $-1.53$ \\
\bottomrule
\end{tabular}
\end{table}

Under the 50\% data regime, all three structural methods outperform the CE baseline by approximately $+$2.3~pp, demonstrating improved sample efficiency. Notably, AuxCrossing-Ord achieves $96.60 \pm 0.97\%$ at 50\% data, matching TACA's gain while exhibiting more consistent behavior across data regimes. The random-distance control further reveals that TACA's benefit under limited data is attributable to general regularization rather than topological content ($\Delta_{\text{spec}} = +0.07$~pp at 50\% data). AuxCrossing-Ord does not exhibit this real-versus-random sensitivity, and together with the MixStyle composability result (Section~\ref{sec:dg_baselines}), this establishes explicit crossing-number supervision as the more robust structural supervision strategy.

\subsection{Domain Generalization and Composability}
\label{sec:dg_baselines}

We investigate how the proposed methods behave under domain generalization (DG) conditions and whether they compose effectively with DG training strategies. MixStyle \citep{zhou2021mixstyle} improves ResNet-50 from 93.96\% to 95.63\% on the canonical split and from 93.13\% to 94.37\% on held-out SLS, demonstrating meaningful robustness gains for CNN-based architectures. Swin-T performance remains stable with or without MixStyle (97.08\% and 97.50\%, respectively), consistent with its stronger baseline representations. Importantly, the combination of MixStyle and AuxCrossing achieves 97.92\%, demonstrating that per-sample auxiliary supervision composes effectively with DG augmentation. AmpMix yields suboptimal performance in this setting, suggesting that frequency-domain perturbations are less suited to the tightness-stratified shift in Knots-10.

\subsection{Appearance Bias and Robustness}
\label{sec:appearance}

We systematically characterize appearance bias through three complementary analyses: appearance ablation, causal probes, and a cross-domain phone-photo evaluation.

\subsubsection{Appearance Ablation}
\label{sec:appearance_ablation}

We evaluate models on original, grayscale, background-masked, and histogram-equalized test images. Grayscale conversion reduces accuracy by 19.6--29.4 percentage points and background masking by 21.0--33.5 percentage points, quantifying the extent to which models rely on color and contextual cues. Notably, on background-masked images, TACA(real) exceeds the CE baseline by 4.4 percentage points, indicating that topology-guided training yields improved robustness to background removal.

\subsubsection{Causal Probes}
\label{sec:causal_probes}

To more directly assess whether models rely on rope crossing structure or non-rope contextual cues, we generate binary rope masks for all 480 test images using U$^2$-Net \citep{qin2020u2net} and define three probe conditions: \textbf{rope-only} (background replaced with neutral gray), \textbf{background-swapped} (same-class background exchanged while rope pixels remain fixed), and \textbf{flip rate} (fraction of predictions that change under the swap). Four trained models are evaluated: Swin-T (CE, TACA, AuxCrossing) and ResNet-50 (CE).

\begin{figure}[H]
\centering
\includegraphics[width=0.96\linewidth]{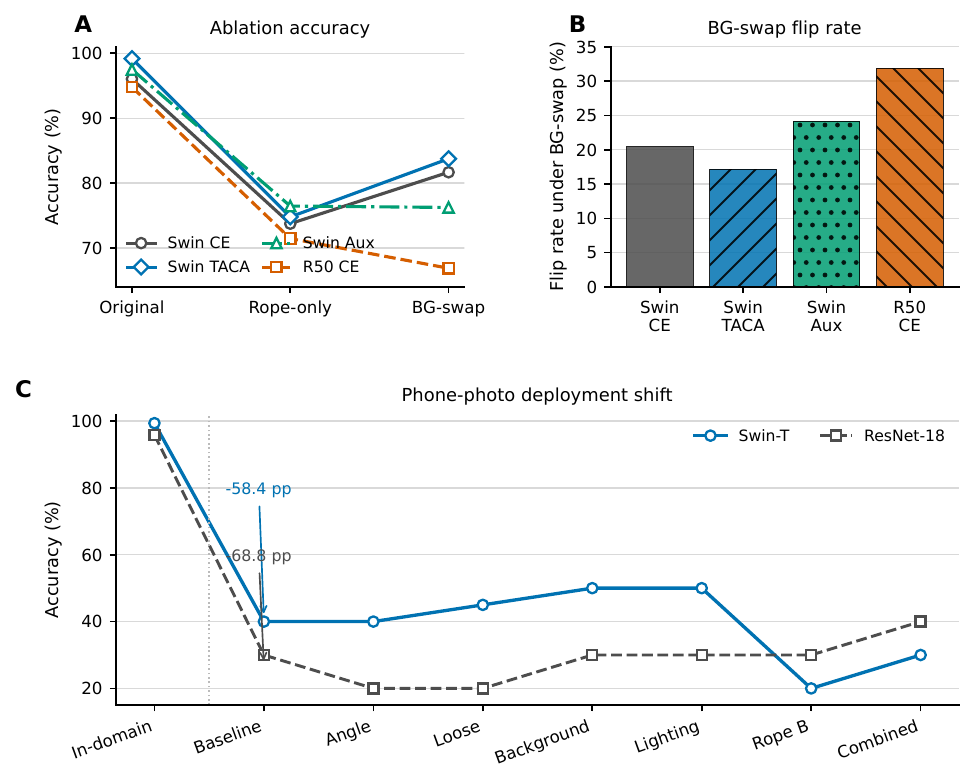}
\caption{\textbf{Appearance bias under causal and deployment shifts.} (A) Accuracy under original, rope-only, and background-swapped conditions for four representative models ($n = 480$). (B) Prediction flip rate under background swapping, where Swin-T with TACA achieves the lowest flip rate among all evaluated models. (C) Accuracy under the phone-photo pilot across progressively shifted capture conditions ($n = 100$), quantifying the deployment gap between controlled and real-world acquisition.}
\label{fig:appearance_summary}
\end{figure}

Two key findings emerge. First, \textbf{Swin-T with TACA achieves the lowest background-swap flip rate and the highest background-swapped accuracy} among all evaluated models (Figure~\ref{fig:appearance_summary}A--B), demonstrating that topology-guided training meaningfully reduces sensitivity to non-structural context. Second, background swapping alone flips 17--32\% of predictions across models, underscoring that appearance bias is a pervasive challenge that warrants targeted mitigation strategies beyond structural supervision alone.

\subsubsection{Cross-Domain Pilot}
\label{sec:crossdomain}

A 100-image cross-domain pilot using smartphone photographs reveals accuracy reductions of 58--69 percentage points (Figure~\ref{fig:appearance_summary}C), with rope material change constituting the dominant source of degradation. These results highlight the substantial domain gap between controlled laboratory acquisition and real-world deployment conditions, and motivate future work on foreground-aware training and stronger appearance augmentation strategies.

\section{Discussion}
\label{sec:discussion}

The central finding of this study is that high closed-set accuracy on Knots-10 does not imply that a model primarily relies on knot topology for its predictions.

Our results demonstrate that topological distance is a meaningful predictor of inter-class confusion, validating the utility of the proposed metric and the Mantel-based screening criterion. Among the evaluated structural supervision strategies, TACA yields consistent topology-specific gains for Swin-T under the canonical protocol ($\Delta_{\text{spec}} = +1.18$~pp), and auxiliary crossing-number prediction offers a more robust and composable alternative across data regimes. Backbone architecture plays an important role: strong transformer-based models benefit more reliably from structural supervision than CNN-based architectures \citep{backbones2024}, suggesting that representational capacity is a prerequisite for topology-specific gains.

Appearance bias remains the principal challenge for deployment. Background swaps flip 17--32\% of predictions, phone-photo accuracy drops by 58--69 percentage points, and standard DG baselines do not close this gap \citep{lrp2024, fgbg2024}. These findings motivate future work on foreground-aware training and stronger background augmentation. The practical implication of our diagnostic workflow is that the Mantel test provides an efficient go/no-go criterion: if the proposed structural metric does not already explain model confusions, distance-guided regularization is unlikely to provide topology-specific benefit.

\section{Conclusion and Limitations}
\label{sec:conclusion}

\textbf{Conclusion.} We introduce Knots-10, a tightness-stratified benchmark and diagnostic testbed for physical knot classification. Our experiments demonstrate that topological distance reliably predicts inter-class confusion patterns across multiple architectures, and that TACA yields consistent topology-specific gains for Swin-T under the canonical protocol. Auxiliary crossing-number prediction provides a more robust and composable structural supervision signal across data regimes. The benchmark's principal contribution is diagnostic: it provides a principled workflow combining Mantel screening, random-distance controls, causal probes, and cross-domain evaluation to rigorously verify whether a hand-crafted structural prior delivers genuine task-relevant benefit beyond generic regularization.

\textbf{Limitations and future work.} The test set is modest in scale, the phone-photo pilot comprises 100 images, and causal probes employ a single checkpoint per model. The deliberately simple training protocol may understate methods that benefit from larger input resolution or extended training schedules. Future work should expand the benchmark with diverse rope materials, backgrounds, and capture devices, and should address appearance bias directly through foreground-aware training, stronger background randomization, or style-based augmentation.


\section*{Declaration of Competing Interest}

The authors declare that they have no known competing financial interests or personal relationships that could have appeared to influence the work reported in this paper.

\section*{CRediT Authorship Contribution Statement}

\textbf{Shiheng Nie}: Methodology, Software, Validation, Formal analysis, Investigation, Data curation, Writing -- original draft, Visualization. \textbf{Yunguang Yue}: Conceptualization, Supervision, Writing -- review \& editing.

\section*{Declaration of Generative AI and AI-Assisted Technologies in the Writing Process}

During the preparation of this work the authors used Gemini (Google) to assist with language polishing. After using this tool, the authors reviewed and edited the content as needed and take full responsibility for the content of the published article.

\section*{Data and Code Availability}

The main benchmark uses the publicly available 10Knots dataset \citep{cameron2018tenknots}, released on Kaggle under a CC BY-SA 4.0 license. The 100 cross-domain phone photographs collected by the authors for the pilot study (Section~\ref{sec:crossdomain}) are available from the corresponding author upon reasonable request. Our evaluation code and analysis scripts are publicly available at \href{https://github.com/Mousaee/knots10-benchmark}{the project repository}.

\section*{Acknowledgements}

This work was inspired by the graduate course \emph{Deep Intelligent Computing and Practice} at Shihezi University. This research received no specific grant from any funding agency in the public, commercial, or not-for-profit sectors.

\bibliographystyle{plain}
\bibliography{references}

\begin{thebibliography}{10}
\expandafter\ifx\csname url\endcsname\relax
  \def\url#1{\texttt{#1}}\fi
\expandafter\ifx\csname urlprefix\endcsname\relax\def\urlprefix{URL }\fi
\expandafter\ifx\csname href\endcsname\relax
  \def\href#1#2{#2} \def\path#1{#1}\fi

\bibitem{wei2022fgia}
X.-S. Wei, Y.-Z. Song, O.~Mac~Aodha, J.~Wu, Y.~Peng, J.~Tang, J.~Yang,
  S.~Belongie, Fine-grained image analysis with deep learning: {A} survey, IEEE
  Transactions on Pattern Analysis and Machine Intelligence 44~(12) (2022)
  8927--8948.
\newblock \href {https://doi.org/10.1109/TPAMI.2021.3126648}
  {\path{doi:10.1109/TPAMI.2021.3126648}}.

\bibitem{adams2004}
C.~C. Adams, The Knot Book: An Elementary Introduction to the Mathematical
  Theory of Knots, American Mathematical Society, 2004.

\bibitem{zimmer2020surgical}
Y.-R. Wong, D.~A. McGrouther, Biomechanics of surgical knot security: a
  systematic review, International Journal of Surgery 109~(3) (2023) 481--490.
\newblock \href {https://doi.org/10.1097/JS9.0000000000000298}
  {\path{doi:10.1097/JS9.0000000000000298}}.

\bibitem{sharma2019complexdna}
R.~K. Sharma, I.~Agrawal, L.~Dai, P.~S. Doyle, S.~Garaj, Complex {DNA} knots
  detected with a nanopore sensor, Nature Communications 10 (2019) 4473.
\newblock \href {https://doi.org/10.1038/s41467-019-12358-4}
  {\path{doi:10.1038/s41467-019-12358-4}}.

\bibitem{hsu2023knottedproteins}
S.-T.~D. Hsu, Folding and functions of knotted proteins, Current Opinion in
  Structural Biology 83 (2023) 102709.
\newblock \href {https://doi.org/10.1016/j.sbi.2023.102709}
  {\path{doi:10.1016/j.sbi.2023.102709}}.

\bibitem{cameron2018tenknots}
J.~Cameron,
  \href{https://www.kaggle.com/datasets/josephcameron/10knots}{{10Knots}: The
  comprehensive dataset of knots}, kaggle dataset, CC BY-SA 4.0 license,
  accessed March 28, 2026 (2018).
\newline\urlprefix\url{https://www.kaggle.com/datasets/josephcameron/10knots}

\bibitem{hughes2020neural}
M.~C. Hughes, A neural network approach to predicting and computing knot
  invariants, Journal of Knot Theory and Its Ramifications 29~(03) (2020)
  2050005.
\newblock \href {https://doi.org/10.1142/S0218216520500054}
  {\path{doi:10.1142/S0218216520500054}}.

\bibitem{vandans2020knots}
O.~Vandans, K.~Yang, Z.~Wu, L.~Dai, Identifying knot types of polymer
  conformations by machine learning, Physical Review E 101~(2) (2020) 022502.
\newblock \href {https://doi.org/10.1103/PhysRevE.101.022502}
  {\path{doi:10.1103/PhysRevE.101.022502}}.

\bibitem{sundaresan2020}
P.~Sundaresan, J.~Grannen, B.~Thananjeyan, A.~Balakrishna, M.~Laskey, K.~Stone,
  J.~E. Gonzalez, K.~Goldberg, Learning rope manipulation policies using dense
  object descriptors trained on synthetic depth, in: Proceedings of the IEEE
  International Conference on Robotics and Automation, 2020, pp. 9411--9418.
\newblock \href {https://doi.org/10.1109/ICRA40945.2020.9197121}
  {\path{doi:10.1109/ICRA40945.2020.9197121}}.

\bibitem{he2022transfg}
J.~He, J.-N. Chen, S.~Liu, A.~Kortylewski, C.~Yang, Y.~Bai, C.~Wang, Trans{FG}:
  A transformer architecture for fine-grained recognition, in: Proceedings of
  the AAAI Conference on Artificial Intelligence, Vol.~36, 2022, pp. 852--860.
\newblock \href {https://doi.org/10.1609/aaai.v36i1.19967}
  {\path{doi:10.1609/aaai.v36i1.19967}}.

\bibitem{wang2023aatrans}
Q.~Wang, J.~Wang, H.~Deng, X.~Wu, Y.~Wang, G.~Hao, {AA-trans}: Core attention
  aggregating transformer with information entropy selector for fine-grained
  visual classification, Pattern Recognition 140 (2023) 109547.
\newblock \href {https://doi.org/10.1016/j.patcog.2023.109547}
  {\path{doi:10.1016/j.patcog.2023.109547}}.

\bibitem{zhuang2020apinet}
P.~Zhuang, Y.~Wang, Y.~Qiao, Learning attentive pairwise interaction for
  fine-grained classification, in: Proceedings of the AAAI Conference on
  Artificial Intelligence, Vol.~34, 2020, pp. 13130--13137.

\bibitem{du2020pmg}
R.~Du, D.~Chang, A.~K. Bhunia, J.~Xie, Z.~Ma, Y.-Z. Song, J.~Guo, Fine-grained
  visual classification via progressive multi-granularity training of jigsaw
  patches, in: Computer Vision -- ECCV 2020, Vol. 12365 of Lecture Notes in
  Computer Science, Springer, 2020, pp. 153--168.
\newblock \href {https://doi.org/10.1007/978-3-030-58565-5_10}
  {\path{doi:10.1007/978-3-030-58565-5_10}}.

\bibitem{chen2024fetfgvc}
H.~Chen, H.~Zhang, C.~Liu, J.~An, Z.~Gao, J.~Qiu, {FET-FGVC}: Feature-enhanced
  transformer for fine-grained visual classification, Pattern Recognition 149
  (2024) 110265.
\newblock \href {https://doi.org/10.1016/j.patcog.2024.110265}
  {\path{doi:10.1016/j.patcog.2024.110265}}.

\bibitem{shi2025ldhvit}
Y.~Shi, Q.~Hong, Y.~Yan, J.~Li, {LDH-ViT}: Fine-grained visual classification
  through local concealment and feature selection, Pattern Recognition 161
  (2025) 111224.
\newblock \href {https://doi.org/10.1016/j.patcog.2024.111224}
  {\path{doi:10.1016/j.patcog.2024.111224}}.

\bibitem{ke2025multigranularity}
X.~Ke, Y.~Cai, B.~Chen, H.~Liu, W.~Guo, Multi-granularity interaction and
  feature recombination network for fine-grained visual classification, Pattern
  Recognition 166 (2025) 111632.
\newblock \href {https://doi.org/10.1016/j.patcog.2025.111632}
  {\path{doi:10.1016/j.patcog.2025.111632}}.

\bibitem{zhou2022domain}
K.~Zhou, Z.~Liu, Y.~Qiao, T.~Xiang, C.~C. Loy, Domain generalization: {A}
  survey, IEEE Transactions on Pattern Analysis and Machine Intelligence 45~(4)
  (2022) 4396--4415.
\newblock \href {https://doi.org/10.1109/TPAMI.2022.3195549}
  {\path{doi:10.1109/TPAMI.2022.3195549}}.

\bibitem{xu2023fourier}
Q.~Xu, R.~Zhang, Z.~Fan, Y.~Wang, Y.-Y. Wu, Y.~Zhang, Fourier-based
  augmentation with applications to domain generalization, Pattern Recognition
  139 (2023) 109474.
\newblock \href {https://doi.org/10.1016/j.patcog.2023.109474}
  {\path{doi:10.1016/j.patcog.2023.109474}}.

\bibitem{hu2024dg_alignment}
J.~Hu, L.~Qi, J.~Zhang, Y.~Shi, Domain generalization via inter-domain
  alignment and intra-domain expansion, Pattern Recognition 146 (2024) 110029.
\newblock \href {https://doi.org/10.1016/j.patcog.2023.110029}
  {\path{doi:10.1016/j.patcog.2023.110029}}.

\bibitem{backbones2024}
S.~Angarano, M.~Martini, F.~Salvetti, V.~Mazzia, M.~Chiaberge, Back-to-bones:
  Rediscovering the role of backbones in domain generalization, Pattern
  Recognition 156 (2024) 110762.
\newblock \href {https://doi.org/10.1016/j.patcog.2024.110762}
  {\path{doi:10.1016/j.patcog.2024.110762}}.

\bibitem{smalldatadg2024}
K.~Chen, E.~Gal, H.~Yan, H.~Li, Domain generalization with small data,
  International Journal of Computer Vision 132 (2024) 3172--3190.
\newblock \href {https://doi.org/10.1007/s11263-024-02028-4}
  {\path{doi:10.1007/s11263-024-02028-4}}.

\bibitem{barz2019hierarchy}
B.~Barz, J.~Denzler, Hierarchy-based image embeddings for semantic image
  retrieval, in: Proceedings of the IEEE Winter Conference on Applications of
  Computer Vision, 2019, pp. 422--431.
\newblock \href {https://doi.org/10.1109/WACV.2019.00073}
  {\path{doi:10.1109/WACV.2019.00073}}.

\bibitem{palazzo2021structured}
S.~Palazzo, F.~Murabito, C.~Pino, F.~Rundo, D.~Giordano, M.~Shah,
  C.~Spampinato, Exploiting structured high-level knowledge for domain-specific
  visual classification, Pattern Recognition 112 (2021) 107806.
\newblock \href {https://doi.org/10.1016/j.patcog.2020.107806}
  {\path{doi:10.1016/j.patcog.2020.107806}}.

\bibitem{he2016resnet}
K.~He, X.~Zhang, S.~Ren, J.~Sun, Deep residual learning for image recognition,
  in: Proceedings of the IEEE Conference on Computer Vision and Pattern
  Recognition, 2016, pp. 770--778.
\newblock \href {https://doi.org/10.1109/CVPR.2016.90}
  {\path{doi:10.1109/CVPR.2016.90}}.

\bibitem{tan2019efficientnet}
M.~Tan, Q.~V. Le, {EfficientNet}: Rethinking model scaling for convolutional
  neural networks, in: Proceedings of the 36th International Conference on
  Machine Learning, 2019, pp. 6105--6114.

\bibitem{dosovitskiy2021vit}
A.~Dosovitskiy, L.~Beyer, A.~Kolesnikov, D.~Weissenborn, X.~Zhai,
  T.~Unterthiner, M.~Dehghani, M.~Minderer, G.~Heigold, S.~Gelly, J.~Uszkoreit,
  N.~Houlsby, An image is worth 16x16 words: Transformers for image recognition
  at scale, in: International Conference on Learning Representations, 2021.

\bibitem{liu2021swin}
Z.~Liu, Y.~Lin, Y.~Cao, H.~Hu, Y.~Wei, Z.~Zhang, S.~Lin, B.~Guo, {Swin
  Transformer}: Hierarchical vision transformer using shifted windows, in:
  Proceedings of the IEEE/CVF International Conference on Computer Vision,
  2021, pp. 10012--10022.
\newblock \href {https://doi.org/10.1109/ICCV48922.2021.00986}
  {\path{doi:10.1109/ICCV48922.2021.00986}}.

\bibitem{kipf2017gcn}
T.~N. Kipf, M.~Welling, Semi-supervised classification with graph convolutional
  networks, in: International Conference on Learning Representations, 2017.

\bibitem{russakovsky2015imagenet}
O.~Russakovsky, J.~Deng, H.~Su, J.~Krause, S.~Satheesh, S.~Ma, Z.~Huang,
  A.~Karpathy, A.~Khosla, M.~Bernstein, A.~C. Berg, L.~Fei-Fei, {ImageNet}
  large scale visual recognition challenge, International Journal of Computer
  Vision 115~(3) (2015) 211--252.
\newblock \href {https://doi.org/10.1007/s11263-015-0816-y}
  {\path{doi:10.1007/s11263-015-0816-y}}.

\bibitem{loshchilov2019adamw}
I.~Loshchilov, F.~Hutter, Decoupled weight decay regularization, in:
  International Conference on Learning Representations, 2019.

\bibitem{zhou2021mixstyle}
K.~Zhou, Y.~Yang, Y.~Qiao, T.~Xiang, Domain generalization with {MixStyle}, in:
  International Conference on Learning Representations, 2021.

\bibitem{wang2021ampmix}
Z.~Wang, Y.~Luo, R.~Qiu, Z.~Huang, M.~Baktashmotlagh, Learning to diversify for
  single domain generalization, in: Proceedings of the IEEE/CVF International
  Conference on Computer Vision, 2021, pp. 834--843.
\newblock \href {https://doi.org/10.1109/ICCV48922.2021.00089}
  {\path{doi:10.1109/ICCV48922.2021.00089}}.

\bibitem{mantel1967}
N.~Mantel, The detection of disease clustering and a generalized regression
  approach, Cancer Research 27~(2) (1967) 209--220.

\bibitem{mcnemar1947}
Q.~McNemar, Note on the sampling error of the difference between correlated
  proportions or percentages, Psychometrika 12~(2) (1947) 153--157.

\bibitem{lrp2024}
P.~R. A.~S. Bassi, S.~S.~J. Dertkigil, A.~Cavalli, Improving deep neural
  network generalization and robustness to background bias via layer-wise
  relevance propagation optimization, Nature Communications 15~(1) (2024) 291.
\newblock \href {https://doi.org/10.1038/s41467-023-44371-z}
  {\path{doi:10.1038/s41467-023-44371-z}}.

\bibitem{fgbg2024}
Q.~Chen, L.~Jiao, F.~Wang, J.~Du, H.~Liu, X.~Wang, R.~Wang, Integrating
  foreground--background feature distillation and contrastive feature learning
  for ultra-fine-grained visual classification, Pattern Recognition 150 (2024)
  110339.
\newblock \href {https://doi.org/10.1016/j.patcog.2024.110339}
  {\path{doi:10.1016/j.patcog.2024.110339}}.

\end{thebibliography}


\begin{thebibliography}{1}
\expandafter\ifx\csname url\endcsname\relax
  \def\url#1{\texttt{#1}}\fi
\expandafter\ifx\csname urlprefix\endcsname\relax\def\urlprefix{URL }\fi
\expandafter\ifx\csname href\endcsname\relax
  \def\href#1#2{#2} \def\path#1{#1}\fi

\bibitem{knotinfo2025}
C.~Livingston, A.~H. Moore, \href{https://knotinfo.org}{{KnotInfo}: Table of
  knot invariants}, accessed March 28, 2026 (2025).
\newline\urlprefix\url{https://knotinfo.org}

\bibitem{wah2011cub}
C.~Wah, S.~Branson, P.~Welinder, P.~Perona, S.~Belongie, The caltech-ucsd
  birds-200-2011 dataset, Tech. Rep. CNS-TR-2011-001, California Institute of
  Technology (2011).

\bibitem{selvaraju2017gradcam}
R.~R. Selvaraju, M.~Cogswell, A.~Das, R.~Vedantam, D.~Parikh, D.~Batra,
  Grad-{CAM}: Visual explanations from deep networks via gradient-based
  localization, in: Proceedings of the IEEE International Conference on
  Computer Vision (ICCV), 2017, pp. 618--626.
\newblock \href {https://doi.org/10.1109/ICCV.2017.74}
  {\path{doi:10.1109/ICCV.2017.74}}.

\bibitem{chefer2021transformer}
H.~Chefer, S.~Gur, L.~Wolf, Transformer interpretability beyond attention
  visualization, in: Proceedings of the IEEE/CVF Conference on Computer Vision
  and Pattern Recognition (CVPR), 2021, pp. 782--791.
\newblock \href {https://doi.org/10.1109/CVPR46437.2021.00084}
  {\path{doi:10.1109/CVPR46437.2021.00084}}.

\end{thebibliography}

\end{document}


\maketitle

\section*{S1. Multi-Seed Robustness Analysis}\label{sec:supp_robustness}

All main-paper results are reported as mean $\pm$ standard deviation across three random seeds (42, 123, 456). Each setting was trained from scratch with the same hyperparameters. Only the random seed changed the weight initialization, data shuffling, and train/validation split. Table~\ref{tab:robustness} gives the test accuracy and macro F1 for each setting across the three runs.

\begin{table}[htbp]
\centering
\caption{Multi-seed robustness analysis. Each configuration is trained with seeds 42, 123, and 456 under the same protocol and evaluated on the tightness-stratified test set.}
\label{tab:robustness}
\small
\begin{tabular}{@{}lcc@{}}
\toprule
Configuration & Test Accuracy (\%) & Macro F1 \\
\midrule
ResNet-18 (CE)             & $96.88 \pm 1.06$ & $0.969 \pm 0.011$ \\
ResNet-18 (CE + TACA)      & $96.81 \pm 1.00$ & $0.968 \pm 0.010$ \\
ResNet-18 (CE + TAML)      & $96.53 \pm 0.79$ & $0.965 \pm 0.008$ \\
ResNet-18 (CE + TACA + TAML) & $95.97 \pm 1.47$ & $0.960 \pm 0.014$ \\
\midrule
ResNet-50 (CE)             & $95.83 \pm 1.03$ & $0.959 \pm 0.010$ \\
EfficientNet-B0 (CE)       & $96.25 \pm 0.45$ & $0.963 \pm 0.005$ \\
ViT-B/16 (CE)              & $96.39 \pm 0.35$ & $0.964 \pm 0.003$ \\
\textbf{Swin-T (CE)}       & $\mathbf{97.22 \pm 1.09}$ & $\mathbf{0.972 \pm 0.011}$ \\
\midrule
\multicolumn{3}{@{}l}{\textit{FGVC-Specialized Methods}} \\
\addlinespace[2pt]
TransFG (CE)               & $97.15 \pm 0.94$ & $0.972 \pm 0.010$ \\
PMG (CE)                   & $94.51 \pm 1.75$ & $0.945 \pm 0.018$ \\
Graph-FGVC (CE)            & $95.49 \pm 2.84$ & $0.955 \pm 0.028$ \\
\midrule
\multicolumn{3}{@{}l}{\textit{Topology-Guided Variants}} \\
\addlinespace[2pt]
ResNet-18 (CE + TACA learnable) & $95.28 \pm 2.23$ & $0.953 \pm 0.022$ \\
ResNet-50 (CE + TACA learnable) & $96.53 \pm 0.43$ & $0.966 \pm 0.004$ \\
\bottomrule
\end{tabular}
\end{table}

Table~\ref{tab:loss_ablation} presents the loss component ablation on ResNet-18, comparing CE only, CE+TACA, CE+TAML, and the combined CE+TACA+TAML configuration.

\begin{table}[htbp]
\centering
\caption{Loss-component ablation on ResNet-18 (mean $\pm$ std across seeds 42, 123, and 456).}
\label{tab:loss_ablation}
\small
\begin{tabular}{@{}lcc@{}}
\toprule
Loss Configuration & Test Acc (\%) & Macro F1 \\
\midrule
CE only & $96.88 \pm 1.06$ & $0.969 \pm 0.011$ \\
CE + TACA & $96.81 \pm 1.00$ & $0.968 \pm 0.010$ \\
CE + TAML & $96.53 \pm \mathbf{0.79}$ & $0.965 \pm 0.008$ \\
CE + TACA + TAML & $95.97 \pm 1.47$ & $0.960 \pm 0.014$ \\
\bottomrule
\end{tabular}
\end{table}

Most settings are stable (std $< 1.5\%$). The main exceptions are Graph-FGVC (2.84\%), learnable-weight ResNet-18 (2.23\%), and PMG (1.75\%). Among the general architectures, the overall ranking remains Swin-T $>$ ResNet-18 $\approx$ ViT-B/16 $\approx$ EfficientNet-B0 $>$ ResNet-50. The FGVC ranking is less stable and changes across seeds, so single-seed comparisons can be misleading. Learnable weights also depend on the backbone: ResNet-18 shows the highest variance ($95.28 \pm 2.23\%$), while ResNet-50 gives the most stable result among the topology-guided settings ($96.53 \pm 0.43\%$).

\emph{Note on cross-table comparisons:} Table~\ref{tab:robustness} reports multi-seed means for the configurations listed here. Some later supplementary analyses use seed-42 values for consistency with embedding or cross-dataset checks. Those single-seed values should not be compared numerically with the multi-seed means in Table~2 of the main text.

\section*{S2. Cross-Domain Evaluation with Phone Photographs}\label{sec:supp_crossdomain}

As a preliminary check on cross-domain generalization, we collected 100 smartphone photographs (10 knot types $\times$ 10 images) under non-laboratory conditions. No calibration or lighting standardization was applied beyond the protocol described below. This was meant to mimic practical deployment. The training set uses blue synthetic climbing rope on a dark background under controlled studio lighting. The cross-domain images use two different ropes (white-red braided nylon cord and black braided cotton cord), three backgrounds (light gray, blue, and beige), two lighting conditions (indoor artificial light and natural window light), and multiple viewing angles (overhead, 45\textdegree, and 30\textdegree). The shift therefore changes rope color, rope material, rope diameter, and background appearance at the same time. These factors are partly confounded and cannot be fully separated. For knot types that require two rope strands (Clove Hitch, Reef Knot, Fisherman's Knot, Flemish Bend), a second gray braided nylon cord was used as the paired strand. Each photograph changes one or more factors relative to the baseline condition in Table~\ref{tab:crossdomain_protocol}.

\begin{table}[htbp]
\centering
\caption{Cross-domain photography protocol. Each knot type has 10 photographs with controlled changes relative to the training distribution. Photo~1 is the closest to the training setup.}
\label{tab:crossdomain_protocol}
\small
\resizebox{\linewidth}{!}{%
\begin{tabular}{@{}clcccc@{}}
\toprule
Photo & Factor Tested & Tightness & Angle & Background & Rope \\
\midrule
1 & Baseline & Tight & 90\textdegree & Gray & A \\
2--3 & Viewing angle & Tight & 45\textdegree/30\textdegree & Gray & A \\
4--5 & Looseness ($\pm$ angle) & Loose & 90\textdegree/45\textdegree & Gray & A \\
6--7 & Background & Tight & 90\textdegree & Blue/Beige & A \\
8 & Lighting & Tight & 90\textdegree & Gray & A$^{\dagger}$ \\
9 & Rope material & Tight & 90\textdegree & Gray & B \\
10 & Combined shift & Tight & 90\textdegree & Blue & B$^{\dagger}$ \\
\bottomrule
\end{tabular}%
}
\vspace{0.3em}

\raggedright\footnotesize $^{\dagger}$Natural light; all other photos use indoor artificial light. Rope A: white-red braided nylon cord ($\sim$6~mm); Rope B: black braided cotton cord ($\sim$4~mm). Both differ substantially from the training rope (blue synthetic climbing rope, $\sim$10~mm) in color, material, and diameter.
\end{table}

\subsubsection*{Overall Results and Per-Factor Analysis}

Overall cross-domain performance is summarized in the main text (\S4.6 \emph{Appearance Bias and Robustness}). The controlled photography protocol used for this pilot study is given in Table~\ref{tab:crossdomain_protocol}.

\subsubsection*{Per-Class Analysis}

Per-class accuracy reveals that certain knot types are more robust to domain shift than others. For Swin-T, Overhand Knot (OHK) achieves 90\% cross-domain recall, likely due to its simple and distinctive visual structure. Clove Hitch (CH) maintains 80\% recall, in line with its unique two-loop-around-pole appearance. In contrast, Bowline (BK) and Flemish Bend (FMB) drop to 0\%, so their diagnostic features apparently depend on the specific rope appearance in the training data.

\subsubsection*{Interpretation}

Most of the accuracy drop appears to come from rope appearance. The baseline condition (Photo~1) already changes three variables at once relative to the training data: rope color (blue $\to$ white-red), material (synthetic climbing rope $\to$ braided nylon), and diameter ($\sim$10~mm $\to$ $\sim$6~mm). Because these factors are confounded, the 40\% baseline accuracy cannot be attributed to color alone. Switching to rope~B (black braided cotton, $\sim$4~mm) causes another $-20$~pp drop. In those images, the crossing structure is hard to resolve, especially for compact knots such as Overhand Knot and Figure-8 Knot. This likely reflects partial \emph{information loss}: the key visual signal is weaker in the image itself, not only shifted in distribution. Background and lighting changes are much smaller ($+10$~pp or neutral), which suggests some invariance from ImageNet pretraining. Although both models degrade strongly, Swin-T (41.0\%) still keeps a 14-point advantage over ResNet-18 (27.0\%). The high cross-domain recall for Clove Hitch (80\% for Swin-T) also matches the shortcut seen in the in-domain Grad-CAM analysis (main text, \S4.2 \emph{Topological Correlation}). Its cross-shaped layout around a post is preserved even when rope appearance changes, so the model may partly rely on scene geometry rather than crossing topology for this class.

\noindent\textbf{Caveat.} This evaluation was conducted by one photographer using consumer smartphones under non-laboratory conditions, with no calibration beyond the protocol in Table~\ref{tab:crossdomain_protocol}. The sample is small (100 images, 10 per class), and each per-factor estimate is based on only 10--20 images. Differences below about 20~pp are therefore not reliable. This setup intentionally mimics deployment, but it limits how strongly we can interpret per-factor effects. A stronger cross-domain benchmark would need multiple photographers, independent control of rope color, material, and diameter, and many more images.

\section*{S3. Supplementary Figures}\label{sec:supp_figures}

The following figures are referenced in the main text and provide supporting detail for the analyses presented therein.

\begin{figure}[htbp]
\centering
\includegraphics[width=\linewidth]{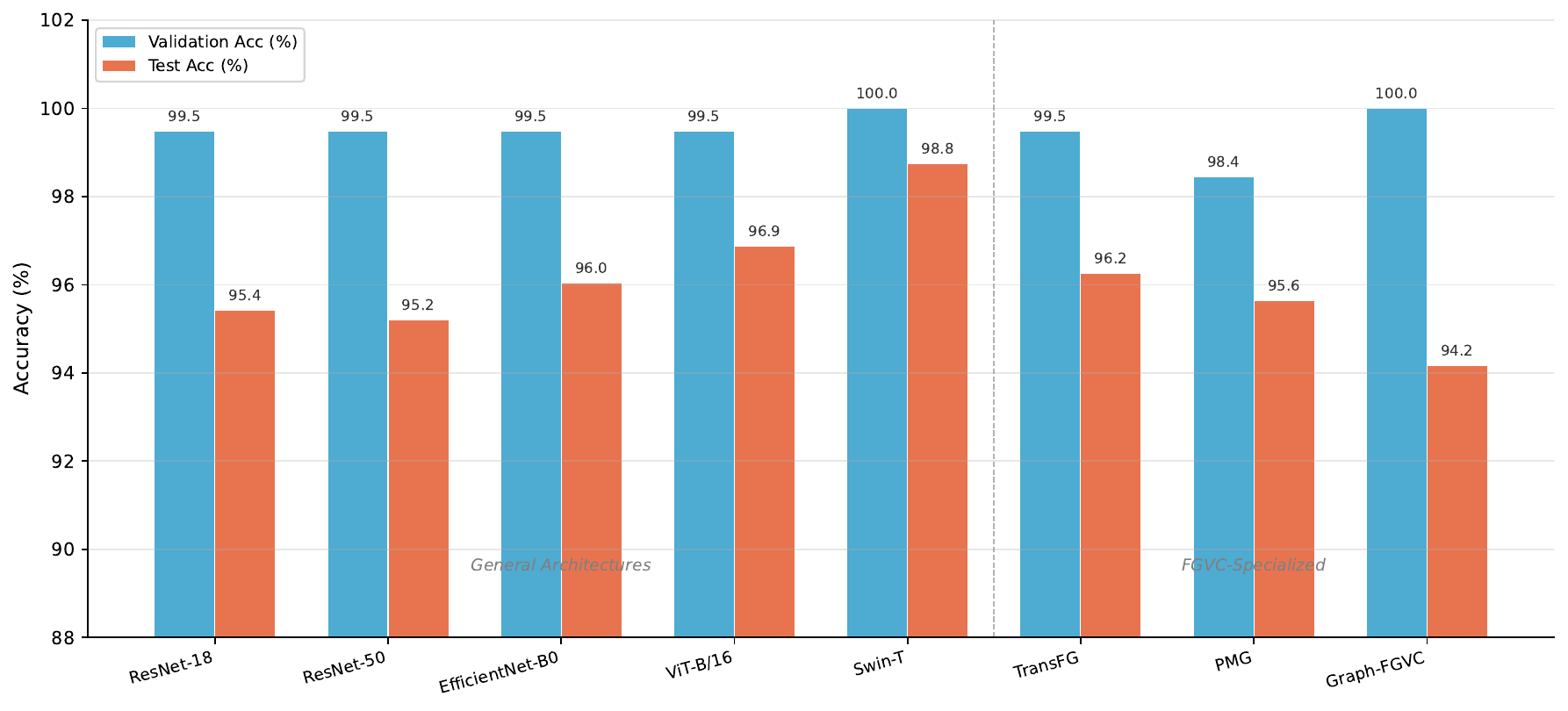}
\caption{Validation and test accuracy for all eight architectures, seed 42. Models are grouped into general-purpose and FGVC-specialized methods. See Table~\ref{tab:robustness} for multi-seed results.}
\label{fig:comparison}
\end{figure}

\begin{figure}[htbp]
\centering
\includegraphics[width=\linewidth]{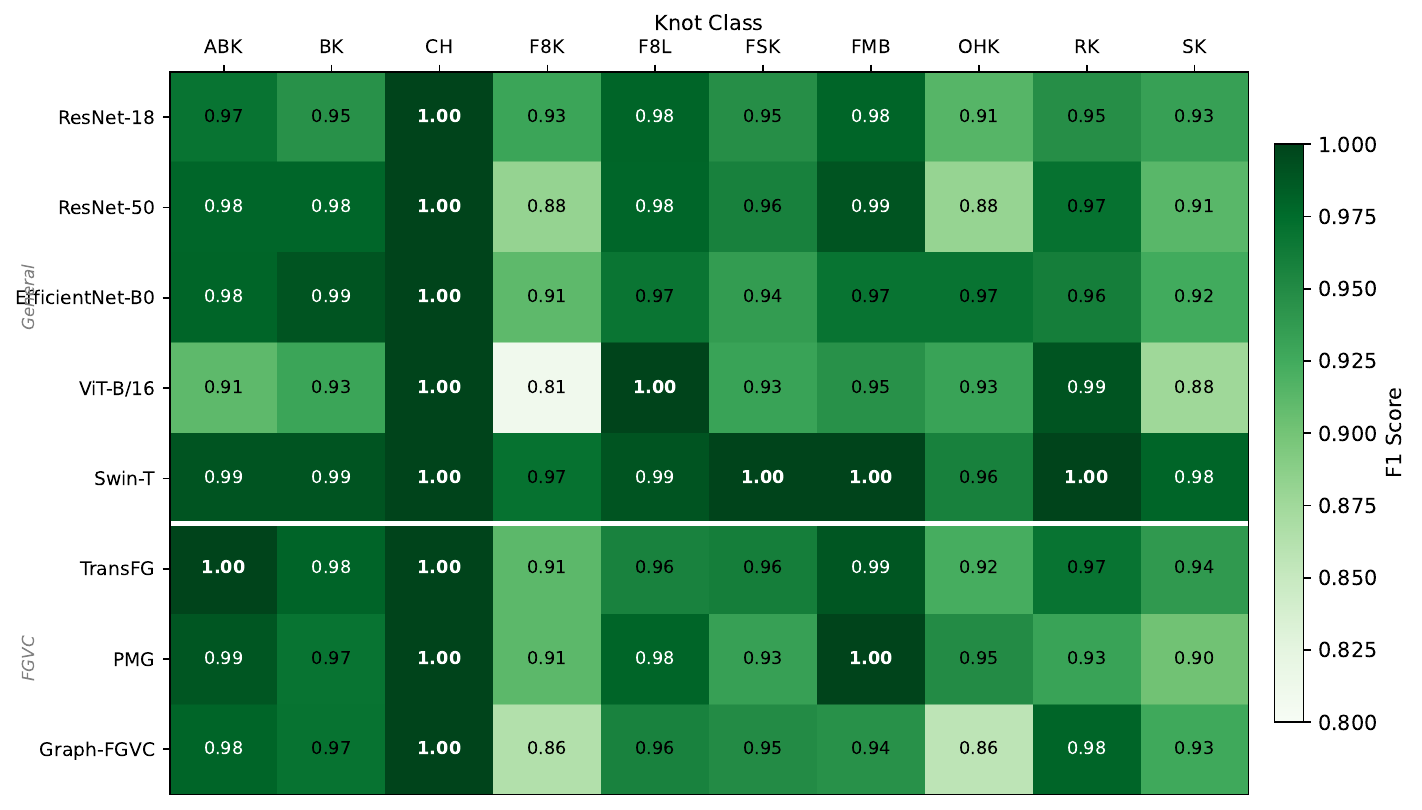}
\caption{Per-class F1 scores for all eight architectures, seed 42. Darker green indicates higher F1. The horizontal line separates general-purpose and FGVC-specialized methods.}
\label{fig:f1heatmap}
\end{figure}

\begin{figure}[htbp]
\centering
\includegraphics[width=\linewidth]{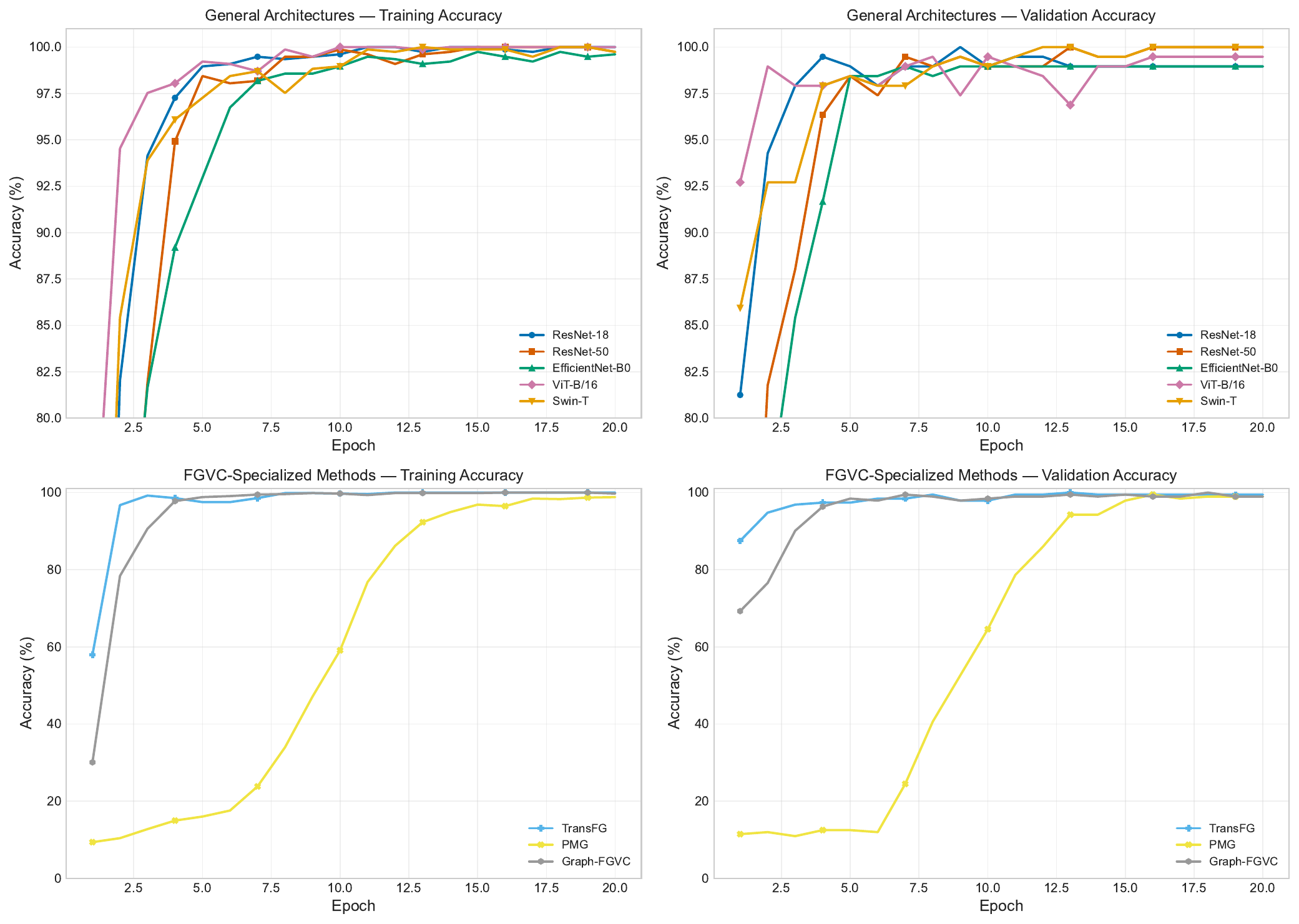}
\caption{Training and validation accuracy curves for all eight architectures over 20 epochs. The top row shows general-purpose models and the bottom row FGVC-specialized methods.}
\label{fig:convergence}
\end{figure}

\begin{figure}[htbp]
\centering
\includegraphics[width=\linewidth]{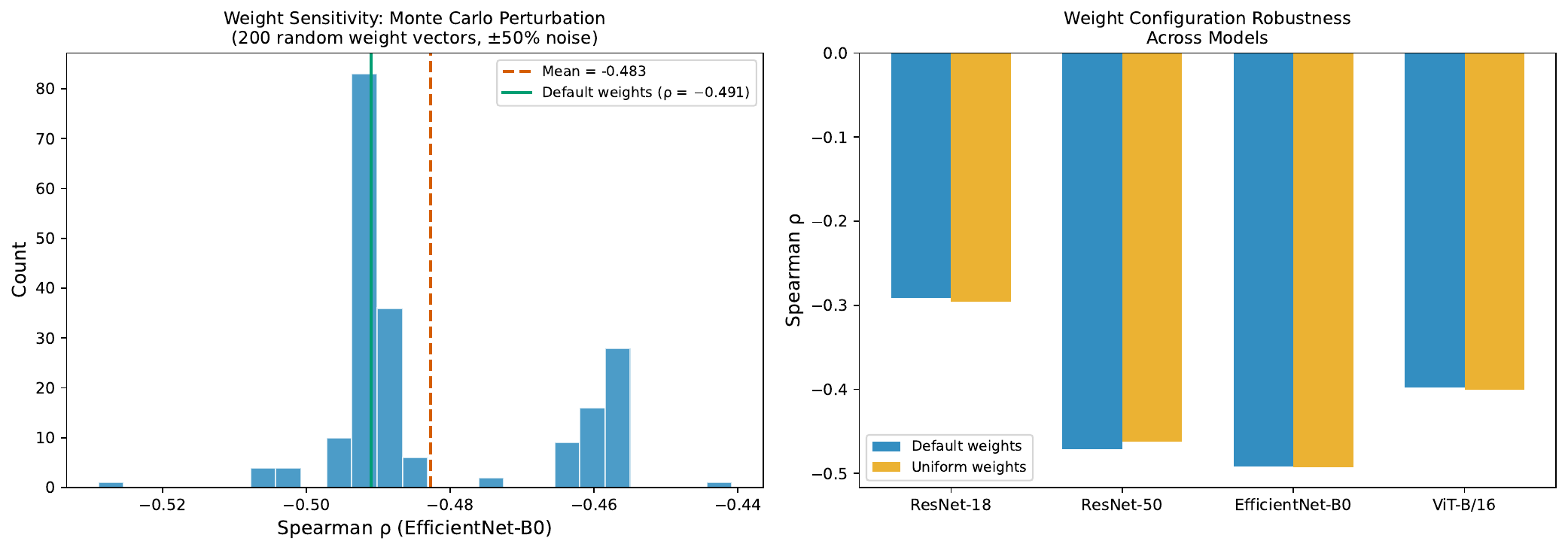}
\caption{Weight sensitivity analysis. Left: Spearman $\rho$ across 200 random weight perturbations ($\pm 50\%$) for EfficientNet-B0. Right: default versus uniform weights across the five general-purpose architectures.}
\label{fig:weight_sensitivity}
\end{figure}

\begin{figure}[htbp]
\centering
\includegraphics[width=\linewidth]{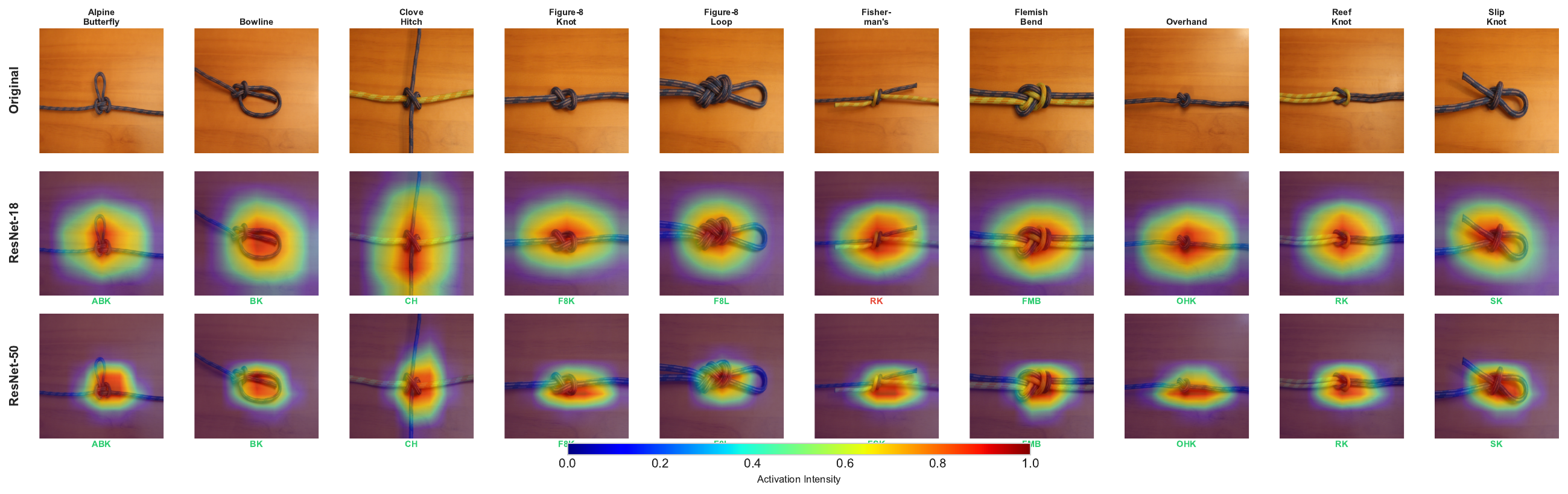}
\caption{Grad-CAM visualizations for ResNet-18 (middle row) and ResNet-50 (bottom row) across all 10 knot classes. The top row shows the input images. Green labels mark correct predictions and red labels mark errors.}
\label{fig:gradcam}
\end{figure}

\begin{figure}[htbp]
\centering
\includegraphics[width=\linewidth]{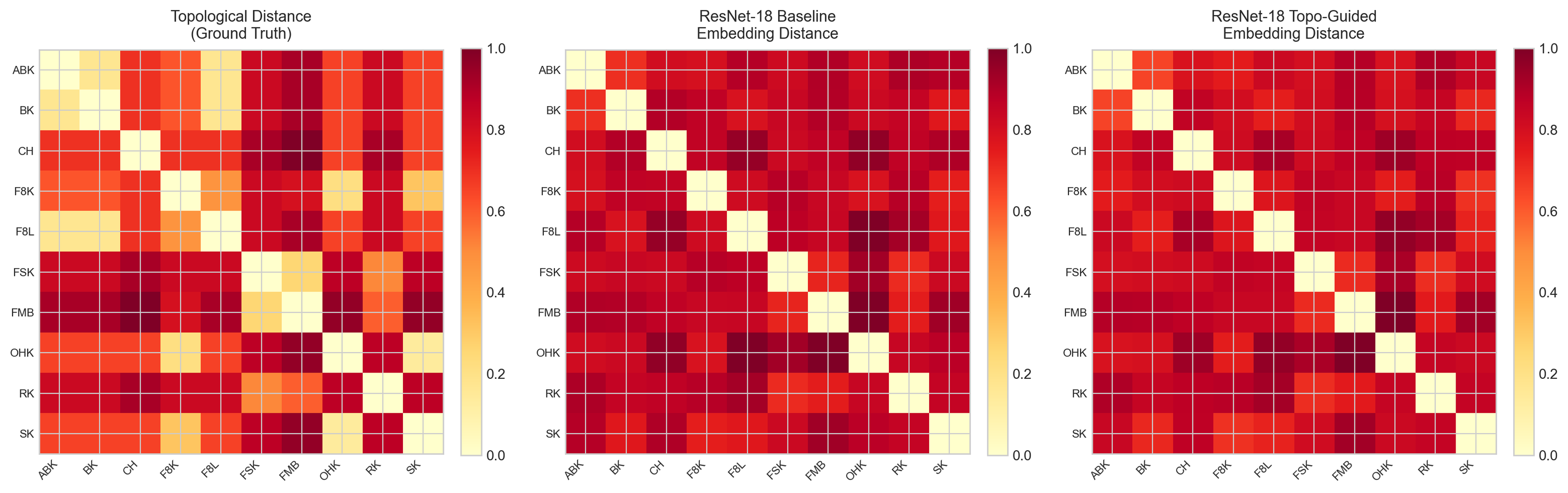}
\caption{Distance matrices for ResNet-18. Left: topological distance. Center: baseline embedding distance. Right: TACA embedding distance. Darker colors indicate larger distances.}
\label{fig:distance_comparison}
\end{figure}

\begin{table}[htbp]
\centering
\caption{Complete $\delta_5$ (structural derivation) matrix for all 45 knot pairs. Four pairs receive reduced penalties; all others use 0.5. Together with Table~1 in the main text, this specifies the full $10 \times 10$ topological distance matrix.}
\label{tab:delta5}
\small
\resizebox{\linewidth}{!}{%
\begin{tabular}{@{}lcccccccccc@{}}
\toprule
 & OHK & SK & F8K & BK & F8L & ABK & CH & RK & FSK & FMB \\
\midrule
OHK & --- & 0.10 & 0.50 & 0.50 & 0.50 & 0.50 & 0.50 & 0.50 & 0.50 & 0.50 \\
SK  &     & --- & 0.50 & 0.50 & 0.50 & 0.50 & 0.50 & 0.50 & 0.50 & 0.50 \\
F8K &     &     & --- & 0.50 & 0.10 & 0.50 & 0.50 & 0.50 & 0.50 & 0.15 \\
BK  &     &     &     & --- & 0.50 & 0.50 & 0.50 & 0.50 & 0.50 & 0.50 \\
F8L &     &     &     &     & --- & 0.50 & 0.50 & 0.50 & 0.50 & 0.50 \\
ABK &     &     &     &     &     & --- & 0.50 & 0.50 & 0.50 & 0.50 \\
CH  &     &     &     &     &     &     & --- & 0.50 & 0.50 & 0.50 \\
RK  &     &     &     &     &     &     &     & --- & 0.10 & 0.50 \\
FSK &     &     &     &     &     &     &     &     & --- & 0.50 \\
FMB &     &     &     &     &     &     &     &     &     & --- \\
\bottomrule
\end{tabular}%
}
\end{table}

\section*{S4. Learnable Weight Evolution}\label{sec:supp_weights}

\begin{figure}[htbp]
\centering
\includegraphics[width=0.7\linewidth]{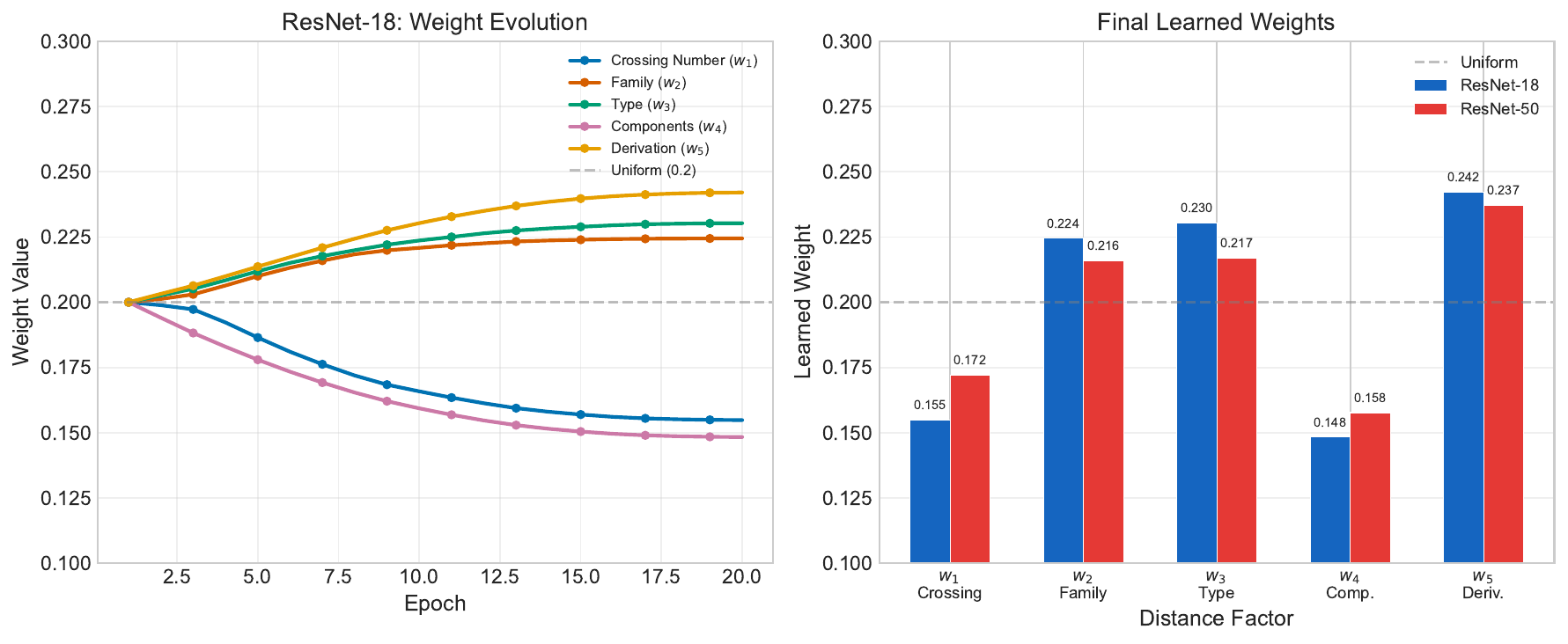}
\caption{Evolution of the five learnable topology weights for ResNet-18. All weights start at 0.2 and stabilize within about 10 epochs.}
\label{fig:weight_evolution}
\end{figure}

Figure~\ref{fig:weight_evolution} shows how the five learnable distance weights change across training epochs for ResNet-18. All weights start at the uniform value $w_f = 0.2$ (equivalently, logits $\ell_f = 0$). During training, the weights move away from uniformity and settle into a stable configuration by about epoch 10.

The final ResNet-18 weights are $\mathbf{w}^*_{18} = (0.157, 0.225, 0.229, 0.153, 0.237)$. This pattern is easy to read. The model gives more weight to structural derivation ($w_5$, from 0.20 to 0.24) and type similarity ($w_3$, from 0.20 to 0.23), and less weight to crossing number ($w_1$, from 0.20 to 0.16) and component count ($w_4$, from 0.20 to 0.15). This is consistent with the idea that knot type and derivation relations may be more visually useful than raw crossing-number differences.

For ResNet-50, only the initial and final weight vectors are available because the full trajectory was not recorded during the original run. Even so, the final configuration $\mathbf{w}^*_{50} = (0.185, 0.211, 0.212, 0.170, 0.223)$ is close to the ResNet-18 result. This suggests that the learned weighting may reflect task structure rather than only backbone-specific effects.

\section*{S5. Mantel Null Distribution}\label{sec:supp_mantel}

\begin{figure}[htbp]
\centering
\includegraphics[width=\linewidth]{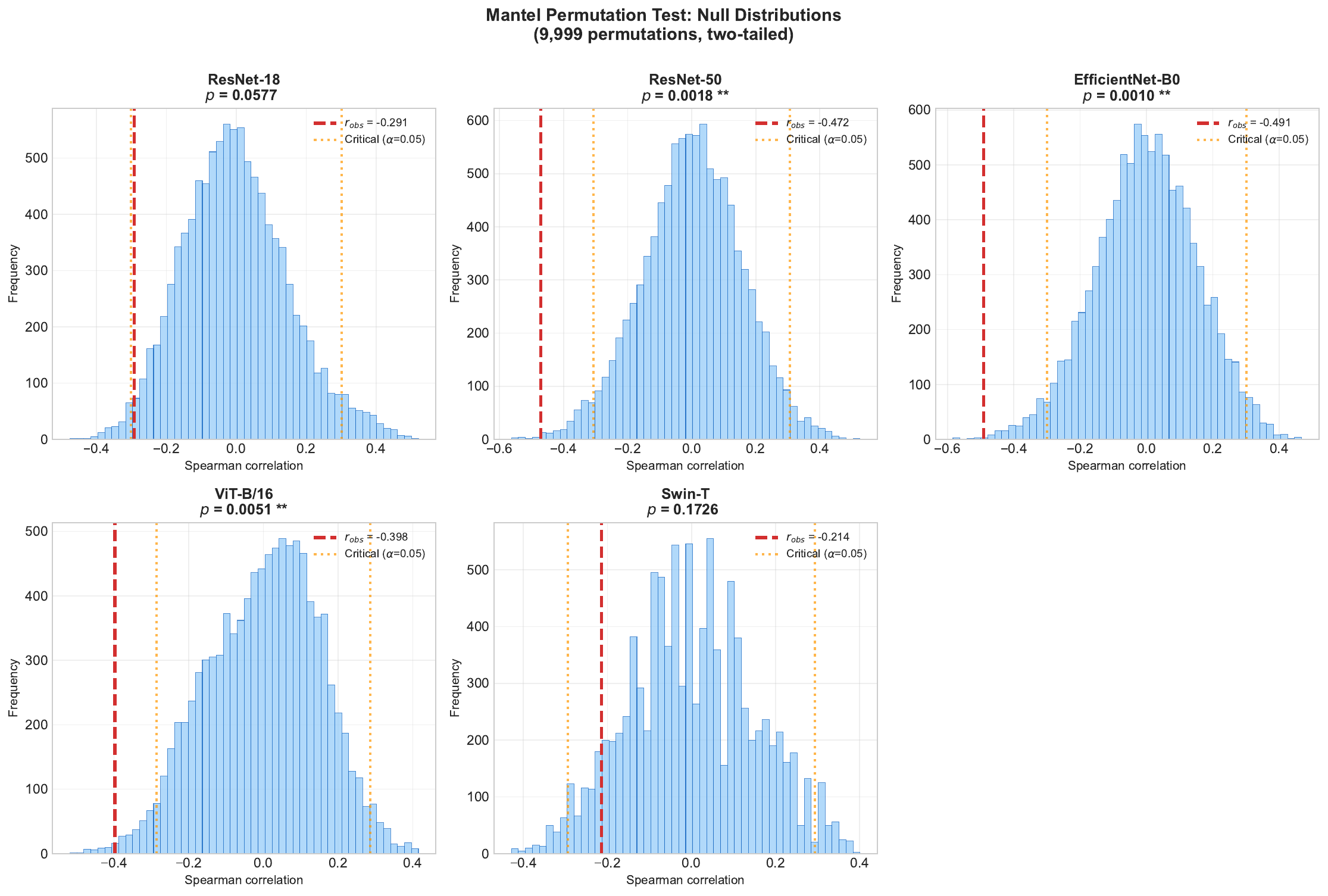}
\caption{Mantel null distributions for the five general-purpose architectures (9{,}999 permutations each). Red dashed lines mark the observed Spearman correlations, and orange dotted lines mark the two-tailed $\alpha = 0.05$ thresholds.}
\label{fig:mantel_null}
\end{figure}

Figure~\ref{fig:mantel_null} shows the null distributions from the Mantel permutation test (9{,}999 random row--column permutations) for all five general-purpose architectures. Each histogram shows the distribution of Spearman correlations under the null hypothesis that topological distance is unrelated to visual confusion rate. The observed correlation (red dashed line) and the two-tailed critical value at $\alpha = 0.05$ (orange dotted lines) are overlaid on each subplot.

The null distributions are approximately symmetric around zero, which suggests that the permutation procedure behaves as expected. The observed correlations for ResNet-50 ($r = -0.472$), EfficientNet-B0 ($r = -0.492$), and ViT-B/16 ($r = -0.398$) fall well into the left tail and clearly pass the significance threshold. Swin-T's observed correlation ($r = -0.214$) stays within the main bulk of the null distribution ($p = 0.173$). The likely reason is simple: Swin-T makes too few errors for a strong correlation test. We therefore read the non-significant result for Swin-T as a power limitation, not as evidence against a topological--visual association.

\section*{S6. McNemar Paired Significance Tests}\label{sec:supp_mcnemar}

\begin{figure}[htbp]
\centering
\includegraphics[width=\linewidth]{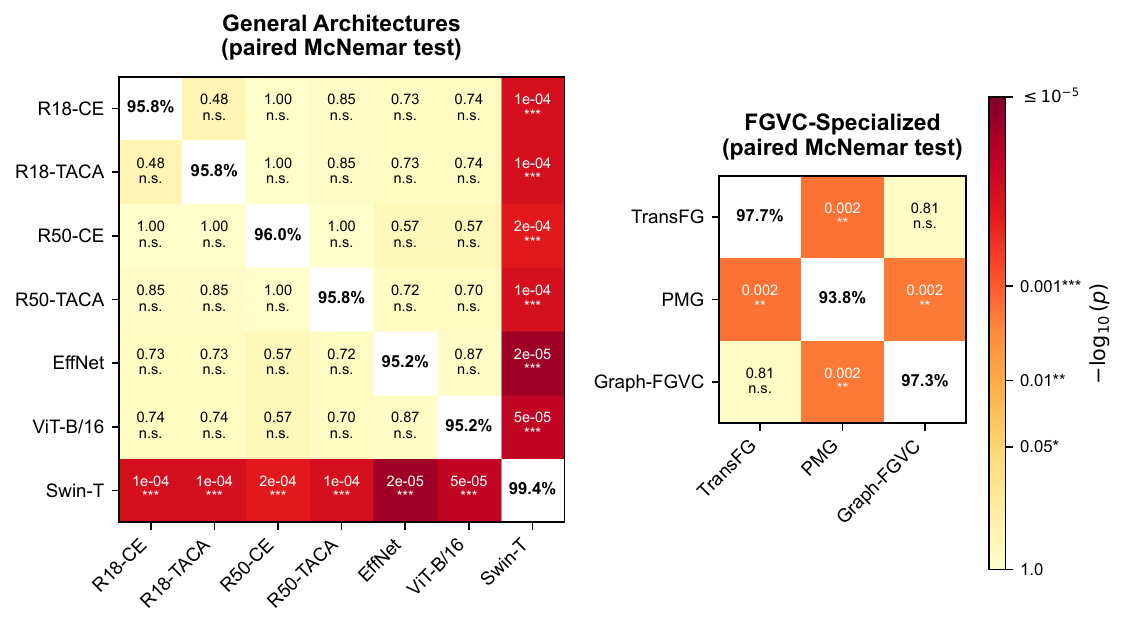}
\caption{McNemar test $p$-values for within-group pairwise comparisons (seed 42, $n = 480$). Color shows $-\log_{10}(p)$. Diagonal cells report test accuracy. Cross-group comparisons are omitted because the test-set order differs.}
\label{fig:mcnemar}
\end{figure}

The McNemar paired significance tests referenced in the main text (\S4.1, \emph{Baseline Results}) are detailed here. Figure~\ref{fig:mcnemar} visualizes the full pairwise $p$-value matrix as a heatmap.

\section*{S7. Embedding Quality Validation: k-NN Retrieval and Mantel Correlation}\label{sec:supp_independent}

This section reports two additional checks for the TACA embeddings analyzed in the main text (Table~4, \S4.3 \emph{Canonical TACA Ablation}).

\textbf{k-NN retrieval accuracy (independent).} For each test sample, we retrieve its $k$ nearest neighbors from the training set embeddings (cosine similarity) and predict the majority-vote label. This metric evaluates whether TACA produces more discriminative embeddings, not merely better-aligned ones, without using the topological distance matrix at evaluation time.

\textbf{Embedding Mantel test (corroborative, not independent).} We compute pairwise Spearman correlation ($\rho$) between the class centroid distance matrix in embedding space and the topological distance matrix, using 9{,}999 permutations. This metric is \emph{not} independent of the training objective (TACA explicitly optimizes for centroid alignment), but it quantifies the magnitude of the effect.

\section*{S8. Random Distance Ablation}\label{sec:supp_ablation}

The random-distance ablation for classification accuracy is reported in the main text (Table~4, \S4.3 \emph{Canonical TACA Ablation}). This section provides additional context on embedding alignment values.

The alignment $\rho$ values in the ablation (0.39--0.45) are lower than those in the k-NN embedding analysis above (Section~S7: 0.47--0.73). This difference likely comes from training initialization and from the fact that the ablation uses a single seed (42) with different random conditions for the permuted matrices. The relative ordering is still consistent (real TACA $>$ permuted $>$ CE), but the size of the gap is sensitive to training details. This is another reason to be cautious about single-seed results.

\section*{S9. Learnable Distance Weights for TACA}\label{sec:supp_learnable}

One limitation of the topology-guided training above is that it uses hand-crafted weights $\mathbf{w} = (0.25, 0.25, 0.15, 0.10, 0.25)$ for the five distance factors. To relax this choice, we extend TACA with \emph{learnable distance weights}. Five scalar logits $\boldsymbol{\ell} \in \mathbb{R}^5$ are passed through a softmax to produce normalized weights $w_f = \exp(\ell_f) / \sum_j \exp(\ell_j)$. These weights are then used to build the topological distance matrix on the fly. The logits are optimized jointly with the model parameters using two learning rates (backbone: $10^{-4}$; weight logits: $10^{-3}$), and they are initialized to uniform weights ($w_f = 0.2$).

Table~\ref{tab:learnable_weights} compares the hand-crafted and learnable weight configurations.

\begin{table}[htbp]
\centering
\caption{Hand-crafted versus learnable topology weights for ResNet-18 and ResNet-50. Test accuracy uses seed 42; multi-seed results are shown where available.}
\label{tab:learnable_weights}
\small
\resizebox{\linewidth}{!}{%
\begin{tabular}{@{}llccc@{}}
\toprule
Backbone & Configuration & Test Acc & Multi-seed Mean & Embed $\rho$ \\
\midrule
\multirow{3}{*}{ResNet-18} & CE only & 95.83\% & $96.88 \pm 1.06$ & 0.462 \\
 & CE + TACA (hand-crafted) & 97.08\% & $96.81 \pm 1.00$ & \textbf{0.645} \\
 & CE + TACA (learnable) & \textbf{96.67\%} & $95.28 \pm 2.23$ & 0.509 \\
\addlinespace
\multirow{3}{*}{ResNet-50} & CE only & 96.04\% & $95.83 \pm 1.03$ & 0.495 \\
 & CE + TACA (hand-crafted) & 95.83\% & ---$^{\dagger}$ & \textbf{0.641} \\
 & CE + TACA (learnable) & \textbf{96.67\%} & $96.53 \pm 0.43$ & 0.505 \\
\bottomrule
\multicolumn{5}{@{}l}{\footnotesize Test Acc column: seed 42 (for consistency with embedding analysis).}\\
\multicolumn{5}{@{}l}{\footnotesize Multi-seed Mean: seeds 42, 123, 456. $^{\dagger}$Not available.}
\end{tabular}
}
\end{table}

Learnable weights behave differently across backbones. ResNet-18 shows the highest variance ($95.28 \pm 2.23\%$, with seed 456 dropping to 92.7\%), while ResNet-50 is much more stable ($96.53 \pm 0.43\%$). Even so, both models converge to similar weight patterns: more weight on structural derivation ($w_5$) and type similarity ($w_3$), and less weight on crossing number ($w_1$) and component count ($w_4$). This suggests that these relations may be more useful for visual discrimination. The weight evolution is shown in Figure~\ref{fig:weight_evolution}.

The embedding alignment ($\rho \approx 0.51$) is lower than hand-crafted TACA ($\rho \approx 0.64$), exposing an accuracy--alignment trade-off: learnable weights optimize for classification rather than explicit distance alignment.

\section*{S10. Validation of Heuristic Distance Against Formal Knot Invariants}\label{sec:supp_knotinfo}

A natural concern about the five-factor heuristic distance (main text, Section~3.2 \emph{Topological Distance}) is that the chosen factors and weights may not match formal knot structure. To partly address this concern, we compare our distance metric with formal knot invariants from the KnotInfo database \citep{knotinfo2025}, which contains rigorously computed properties for more than 12{,}000 mathematical knots.

Among our 10 classes, only four have formal mathematical counterparts: Overhand Knot ($3_1$, the trefoil), Figure-8 Knot ($4_1$), Reef Knot / Fisherman's Knot ($3_1 \# 3_1$), and Flemish Bend ($4_1 \# 4_1$). For the prime knots, we directly extract crossing number, 3-genus, signature, determinant, and Alexander polynomial from KnotInfo. For the composite knots, we derive these invariants using standard connect-sum formulas: crossing number and genus are additive, signature is additive, determinant is multiplicative, and the Alexander polynomial of $K_1 \# K_2$ is the product of the individual polynomials.

We then build an invariant-based distance by normalizing and equally weighting five components: crossing-number difference, genus difference, signature difference, determinant difference, and a coefficient-vector distance for the Alexander polynomial. We compute this distance for the six pairwise comparisons among the four knots. The Spearman rank correlation between the invariant-based distance and our heuristic distance is $\rho = 0.49$ ($p = 0.33$, $n = 6$). The correlation is moderate, but the $p$-value is not significant because the sample is very small. The two metrics agree on the closest pair (Overhand Knot--Figure-8 Knot) and on which knots are most distant. The largest disagreement is Overhand Knot--Flemish Bend, where the invariant-based metric gives a larger distance because of the Alexander-polynomial term.

This analysis has two clear limitations. First, only 4 of our 10 classes have mathematical counterparts, so the remaining 6 cannot be checked in this way. Second, mathematical knot theory is defined on closed curves, whereas our classes are open rope structures, so the correspondence is only approximate. Even with these limits, the partial agreement suggests that the heuristic distance is not arbitrary.

\section*{S11. Cross-Dataset Generalization of TACA}\label{sec:supp_crossdataset}

To test whether topology-guided training is specific to knots, we also apply TACA to CUB-200-2011 \citep{wah2011cub} (200 bird species). We replace the knot-theoretic distance matrix with a taxonomic hierarchy distance (same species: 0, same genus: 0.25, same family: 0.50, same order: 0.75, different order: 1.0). The experiment uses ResNet-50 with the same training protocol (100 epochs, batch size 16, lr=$10^{-4}$, image size 448$\times$448). Table~\ref{tab:cross_dataset} reports the results.

\begin{table}[htbp]
\centering
\caption{Cross-dataset generalization of TACA. ResNet-50 is evaluated on Knots-10 and CUB-200-2011 with dataset-specific distance matrices. For Knots-10, the seed-42 rows are shown for consistency with the cross-dataset comparison; the main paper reports multi-seed results.}
\label{tab:cross_dataset}
\small
\begin{tabular}{@{}llccc@{}}
\toprule
Dataset & Method & Test Acc & Macro F1 & Embed $\rho$ \\
\midrule
\multirow{2}{*}{Knots-10} & CE only & 96.04\% & 0.961 & 0.495 \\
 & CE + TACA & 95.83\% & 0.959 & \textbf{0.641} \\
\addlinespace
\multirow{2}{*}{CUB-200-2011} & CE only & 81.81\% & 0.819 & 0.109 \\
 & CE + TACA & 80.67\% & 0.807 & 0.086 \\
\bottomrule
\end{tabular}
\end{table}

On Knots-10, the seed-42 ResNet-50 run shows no classification gain from TACA ($96.04\% \to 95.83\%$, within sampling noise), but it does improve embedding alignment ($\rho$: $0.495 \to 0.641$, about 30\%). In the multi-seed evaluation in the main paper, however, ResNet-50's topology-specific effect is near zero because TACA(real) and TACA(rand) perform similarly on average. On CUB-200-2011, TACA gives a small accuracy decrease (81.81\% $\to$ 80.67\%) and no improvement in embedding alignment ($\rho$: 0.109 $\to$ 0.086).

The contrast is useful. On knots, the distance matrix captures some visual similarity, as supported by the Mantel test in the main text (\S4.2 \emph{Topological Correlation}). TACA can therefore improve representation alignment even without a clear classification gain. On CUB-200, the taxonomic distance reflects biological relatedness rather than visual similarity. The embedding--taxonomy Spearman correlation on CUB-200 is only $\rho = 0.109$, which shows weak alignment between the taxonomic hierarchy and the learned feature space. This suggests a simple diagnostic workflow. \emph{Before} using structure-guided training, one should first test whether the proposed distance metric correlates with the baseline model's confusion structure, for example with a Mantel permutation test. If that correlation is weak, the distance metric is less likely to help.

\section*{S12. Grad-CAM Attribution Analysis}\label{sec:supp_attribution}

To test whether topology-guided training redirects attention toward the rope region, we computed Grad-CAM \citep{selvaraju2017gradcam} maps for 100 randomly sampled test images. We then measured the fraction of total attribution mass inside the rope segmentation mask (the ``attribution rope ratio''). Rope masks were generated using U$^2$-Net. For Swin-T models, the target layer was the output of the final stage (\texttt{features[-1]}). For ResNet-50, the target layer was the last convolutional block (\texttt{layer4[-1]}).

\begin{table}[htbp]
\centering
\caption{Grad-CAM attribution rope ratio (mean $\pm$ std, $n=100$). The metric is the fraction of attribution mass inside the rope mask. Only within-architecture comparisons are valid.}
\label{tab:attribution}
\small
\begin{tabular}{@{}lc@{}}
\toprule
Model & Attribution rope ratio (\%) \\
\midrule
\multicolumn{2}{@{}l}{\textit{Swin-T variants (within-architecture comparison)}} \\
\addlinespace[2pt]
Swin-T CE & $8.87 \pm 3.24$ \\
Swin-T TACA & $7.96 \pm 3.48$ \\
Swin-T AuxCrossing & $7.72 \pm 3.46$ \\
\midrule
\multicolumn{2}{@{}l}{\textit{CNN baseline}} \\
\addlinespace[2pt]
ResNet-50 CE & $38.30 \pm 12.06$ \\
\bottomrule
\end{tabular}
\end{table}

Within the Swin-T family, all three training conditions produce similar attribution rope ratios (7.7--8.9\%). This suggests that neither TACA nor AuxCrossing redirects attention toward the rope region. The much higher ratio for ResNet-50 (38.3\%) should not be read as stronger rope focus. It more likely reflects architectural differences in gradient flow between CNNs and transformers; Grad-CAM maps in transformers are often more spatially diffuse because of global self-attention \citep{chefer2021transformer}. We therefore avoid cross-architecture conclusions from these raw ratios.

\section*{S13. Reduced-Data Stress Test: Per-Seed Breakdown}\label{sec:supp_reduced_perseed}

Table~\ref{tab:reduced_perseed} reports per-seed results for the reduced-data stress test (main paper, \S4.4 \emph{Reduced-Data Stress Test}). This experiment was conducted as an independent suite with max-normalized TACA distances on both sides, which differs from the main ablation protocol (main paper, Table~4 \emph{Canonical TACA Ablation}); absolute accuracies should not be compared across tables.

\begin{table}[htbp]
\centering
\caption{Per-seed results for the reduced-data stress test on Swin-T. All values are test accuracy (\%) on $n=480$ images.}
\label{tab:reduced_perseed}
\small
\begin{tabular}{@{}lcccccc@{}}
\toprule
 & \multicolumn{3}{c}{50\% data} & \multicolumn{3}{c}{100\% data} \\
\cmidrule(lr){2-4} \cmidrule(lr){5-7}
Method & s42 & s123 & s456 & s42 & s123 & s456 \\
\midrule
CE & 95.00 & 92.50 & 95.42 & 96.04 & 97.71 & 97.71 \\
TACA (real) & 97.29 & 96.67 & 96.04 & 97.08 & 96.67 & 96.25 \\
TACA (rand) & 96.25 & 97.08 & 96.46 & 98.96 & 98.13 & 97.50 \\
AuxCrossing-Ord & 95.83 & 97.71 & 96.25 & 97.50 & 96.67 & 97.08 \\
\bottomrule
\end{tabular}
\end{table}

At 50\% data, all three structural methods show lower variance than CE (std $\leq 0.97$ vs.\ 1.58). This suggests that structural supervision stabilizes training under reduced data. At 100\% data, TACA(rand) outperforms TACA(real) for all three seeds, which rules out a seed-specific artifact.

\section*{S14. Pair-Level Evidence (Seed 42)}\label{sec:supp_pair_bin}

Although the sample-level McNemar test for TACA(real) vs.\ TACA(rand) was not significant, an exploratory pair-level analysis on seed 42 provides a more detailed view. For each of the 45 class pairs, we computed pairwise accuracy under TACA(real) and TACA(rand), and defined $\Delta_{\text{spec}}^{\text{pair}}$ as their difference. We then grouped pairs by topological distance into \emph{near}, \emph{medium}, and \emph{far}. Bootstrap resampling (10{,}000 iterations) was used to build 95\% confidence intervals for the mean $\Delta_{\text{spec}}^{\text{pair}}$ within each bin.

On \textbf{Swin-T}, the bootstrap CI for $\Delta_{\text{spec}}^{\text{pair}}$ did not cross zero in any bin: near pairs $+$1.34~pp [0.74, 2.08], medium $+$1.16~pp [0.69, 1.62], and far $+$0.83~pp [0.52, 1.20]. The gain is largest for near pairs, but it remains positive in all three bins. In the \textbf{seed-42 run of ResNet-50}, the pattern reverses: near pairs $-$0.74~pp [$-$1.93, $+$0.60] (CI crosses zero), medium $-$1.79~pp [$-$3.82, $-$0.23], and far $-$4.22~pp [$-$6.35, $-$2.08]. Because the multi-seed average for ResNet-50 is near zero, we treat this as a single-seed observation rather than a general CNN effect.

\textbf{Mechanistic interpretation.} A simple explanation is that Swin-T's shifted-window self-attention can model longer-range spatial relations and may therefore align better with the global crossing structure encoded in the topological distance matrix. ResNet-50, by contrast, builds its representation from local convolution filters, and the topological distance may not map as directly onto local texture cues. This remains only a possible explanation. The positive $\Delta_{\text{spec}}$ could also come from optimization dynamics rather than representational capacity. As shown in the main paper (\S4.4, \emph{Reduced-Data Stress Test}), the reduced-data experiment gives $\Delta_{\text{spec}} \approx 0$ under a different TACA normalization. So even the positive Swin-T response is sensitive to implementation details.

\section*{S15. Supporting Analyses on CNN Backbones}\label{sec:supp_cnn}

Additional analyses on CNN backbones suggest that topology-specific benefit is weak or unstable on CNNs. On ResNet-18, multi-seed evaluation showed no significant CE vs.\ TACA difference ($p = 0.48$, McNemar; Table~S1), and a four-way loss ablation gave statistically indistinguishable accuracy across all settings (Table~S2). On ResNet-50, the seed-42 embedding and cross-dataset checks in Sections~S9--S11 show that TACA can improve alignment without yielding a stable classification gain. Taken together, these supporting analyses do not provide strong evidence for a consistent CNN benefit from TACA. Learnable distance weights are discussed in Section~S9.

\bibliographystyle{plain}
\bibliography{references}